\documentclass{article}

 \PassOptionsToPackage{numbers, compress}{natbib}

 \usepackage[main, final]{neurips_2026}

\usepackage[utf8]{inputenc} 
\usepackage[T1]{fontenc}    
\usepackage{hyperref}       
\usepackage{url}            
\usepackage{booktabs}       
\usepackage{amsfonts}       
\usepackage{nicefrac}       
\usepackage{microtype}      
\usepackage{xcolor}         
\usepackage{graphicx}
\usepackage{amsmath}

\newtheorem{rmk}{Remark}

\title{FaithfulFaces: Pose-Faithful Facial Identity Preservation for Text-to-Video Generation}

\author{Yuanzhi~Wang$^{1,2,}$\thanks{Work done during the internship at Tencent Hunyuan.}~~,~Xuhua Ren$^{2}$,~Jiaxiang Cheng$^{2}$,~Bing Ma$^{2}$,~Kai Yu$^{2}$,\\
\textbf{Sen Liang$^{2,3}$,~Wenyue Li$^{2}$,~Tianxiang Zheng$^{2}$,~Qinglin Lu$^{2,}$\thanks{Corresponding Authors: Qinglin Lu and Zhen Cui}~~,~Zhen Cui$^{4,}$\footnotemark[2]}\\
\\
	1.~Nanyang Technological University \\
    2.~Tencent Hunyuan\\
 3.~University of Science and Technology of China \\
 4. Beijing Normal University
}

\begin{document}

\maketitle

\begin{abstract}
Identity-preserving text-to-video generation (IPT2V) empowers users to produce diverse and imaginative videos with consistent human facial identity.
Despite recent progress, existing methods often suffer from significant identity distortion under large facial pose variations or facial occlusions.
In this paper, we propose \textit{FaithfulFaces}, a pose-faithful facial identity preservation learning framework to improve IPT2V in complex dynamic scenes.
The key of FaithfulFaces is a pose-shared identity aligner that refines and aligns facial poses across distinct views via a pose-shared dictionary and a pose variation–identity invariance constraint.
By mapping single-view inputs into a global facial pose representation with explicit Euler angle embeddings, FaithfulFaces provides a pose-faithful facial prior that guides generative foundations toward robust identity-preserving generation.
In particular, we develop a specialized pipeline to curate a high-quality video dataset featuring substantial facial pose diversity.
Extensive experiments demonstrate that FaithfulFaces achieves state-of-the-art performance, maintaining superior identity consistency and structural clarity even as pose changes and occlusions occur.
\textbf{Project Page:} \url{https://mdswyz.github.io/FaithfulFaces/} 
\end{abstract}

\section{Introduction}
Identity-preserving text-to-video generation (IPT2V) is a specialized facet of content creation that aims to generate various videos from the user-provided reference image and text prompts while maintaining consistent human facial identity across consecutive frames~\cite{ConsisID,Stand-In}.
This task showcases the potential to create and author visual content across domains, including but not limited to film and television production, personalized avatars, advertising design, and social multimedia content.

Benefiting from the robust generative capabilities of the large-scale pre-trained video foundational generative models~\cite{Hunyuanvideo,CogVideoX,Wan}, the IPT2V task can seamlessly extend these models to generate videos guided by reference face images.
To generate videos with high-fidelity facial identity, researchers have proposed various methods to represent the identity information of the reference image.
For example, ID-Animator~\cite{Id-animator} used a lightweight face adapter to encode the identity-relevant embeddings.
ConsisID~\cite{ConsisID} designed two facial extractors to extract global low-frequency structure and local high-frequency details for IPT2V.
At the same time, many commercial tools, such as Vidu~\cite{Vidu}, Kling~\cite{Kling}, have also been adapted to the IPT2V task.
Therefore, this task is the focus of the GenAI field and has attracted widespread attention.

\begin{figure*}[t]
\centering{\includegraphics[width=0.95\linewidth]{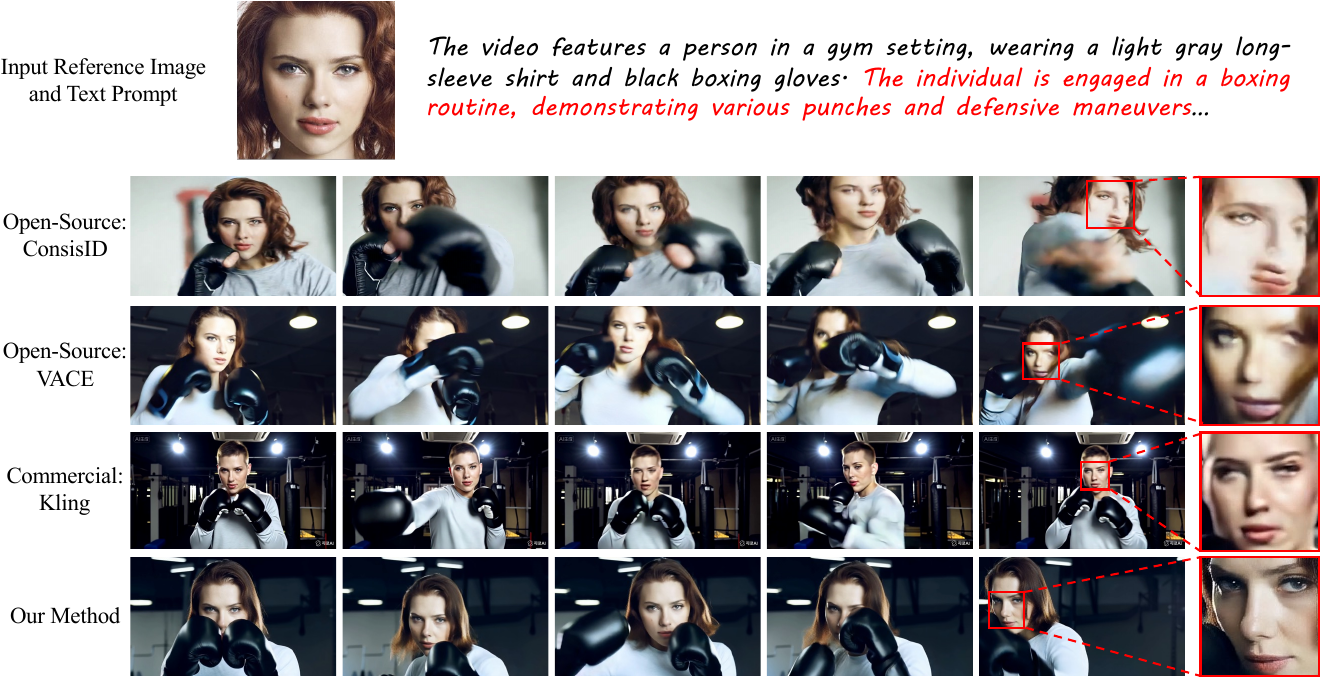}}
	\caption{Visualization results from four different IPT2V methods. ConsisID~\cite{ConsisID} shows severe distortion of the facial structure. VACE~\cite{VACE} and Kling~\cite{Kling} suffer from significant distortion of facial identity details. In contrast to these open-source and commercial methods, our method exhibits clear facial structure and high-fidelity identity details as the facial pose changes and occlusions occur.
	}
	\label{fig:1}
\end{figure*}

Despite their notable success, existing methods still exhibit limitations in effectively handling certain intricate scenarios.
As shown in Fig.~\ref{fig:1}, we visualize the generation results of different methods in a complex dynamic case, where ConsisID and VACE~\cite{VACE} are two representative open-source methods, based on CogVideoX-5B~\cite{CogVideoX} and Wan2.1-14B~\cite{Wan}, respectively. 
Kling is one of the most popular and powerful commercial models.
In this case, the goal is to generate a video depicting a subject performing a boxing action, which often involves significant variations in facial pose as well as facial occlusions.
We can observe that both open-source and commercial approaches tend to produce noticeable distortion in the facial region as the subject moves and their facial expressions or pose change.
This phenomenon may be attributed to the fact that such methods can only capture a single facial pose information from an input reference image, limiting their ability to handle scenarios with significant variations in facial pose.
A question arises: \textit{Can we capture global facial pose information from an input single-view image?}

In this paper, we propose a pose-faithful facial identity preservation learning framework, named \textit{FaithfulFaces}, to address the aforementioned problem.
We first propose a pose-shared identity aligner to encode global facial pose representation from the input single-view reference image.
This aligner establishes a pose-shared dictionary to project diverse facial poses into a refined dictionary space, which is learned by a well-crafted pose variation–identity invariance constraint.
In this constraint, face images from the same identity but with different poses are treated as positive pairs, while others serve as negative samples.
In particular, we incorporate Euler angle embedding learning into the aligner to provide explicit pose cues during the refinement and alignment processes.

Furthermore, to support our FaithfulFaces learning, we design a new dataset collection and processing pipeline that constructs a high-quality, task-specific video dataset with significant facial pose variations to provide a robust training foundation.
Finally, the well-trained framework is capable of naturally extracting global facial pose representations as holistic facial priors, enabling foundational generative models to better preserve identity in generated videos.
As illustrated in Fig.~\ref{fig:1}, our method demonstrates superior consistency in maintaining facial identity throughout the generated video as the facial pose changes and occlusions occur.
The contributions of this work are threefold:
\begin{itemize}
    \item We systematically analyze the limitations and potential reasons of existing IPT2V methods in complex facial dynamic scenes, and propose a pose-faithful facial identity preservation learning paradigm, FaithfulFaces, to better preserve consistent identity in generated videos.
    \item We design a pose-shared identity aligner to encode global facial pose representation from the input single-view reference image via a pose-shared dictionary and a pose variation–identity invariance constraint with Euler angle embedding learning. Additionally, we develop a new dataset pipeline to construct a task-oriented, high-quality video dataset with substantial facial pose diversity to ensure robust model training.
    \item We perform extensive experiments across diverse identity and dynamic scenarios. Both quantitative and qualitative results demonstrate the effectiveness of our FaithfulFaces, surpassing existing open-source and commercial methods.
\end{itemize}

\section{Related Work}
Thanks to the powerful data distribution modeling capability and stable training process of the continuous-time generative models~\cite{scoresde,fm,rf,IMDer,DiCMoR,jin2025latexblend,IMPFR}, large-scale text-to-video generative models~\cite{Moviegen,Hunyuanvideo,CogVideoX,Wan,Seedance,TCVE,jin2026latent,ReAtCo,TraFrCo,IPA} have been rapidly developed, further facilitating the Identity-preserving text-to-video generation (IPT2V) task.
In the early stage, He \textit{et al.}~\cite{Id-animator} proposed the ID-Animator method that uses a Unet-based lightweight text-to-video model AnimateDiff~\cite{Animatediff} and builds a face adapter for IPT2V.

The recent Diffusion Transformer (DiT) architecture~\cite{dit} has shown promising generative capabilities and has become a mainstream backbone for video generation, such as open-source models HunyuanVideo~\cite{Hunyuanvideo}, CogVideoX~\cite{CogVideoX}, and Wan~\cite{Wan}.
Therefore, many recent IPT2V works are built upon and extend the DiT-based models~\cite{ConsisID,Fantasyid,Magicmirror,Stand-In,Concat-ID,Skyreels-a2,MAGREF}.
For example, ConsisID~\cite{ConsisID} utilized CogVideoX as the basic generative model and designed a global and local facial extractor to capture global structure and local details as identity information.
HunyuanCustom~\cite{Hunyuancustom} was built upon the HunyuanVideo foundational model.
VACE~\cite{VACE}, Phantom~\cite{Phantom}, SkyReels-A2~\cite{Skyreels-a2}, MAGREF~\cite{MAGREF}, and Stand-In~\cite{Stand-In} used Wan as the foundational model.

Furthermore, due to the extremely broad range of real-world applications for IPT2V, numerous successful commercial models and tools have emerged, such as Vidu~\cite{Vidu}, Pika~\cite{Pika}, Kling~\cite{Kling}.
However, whether open-source methods or commercial tools, they are difficult to deal with complex facial dynamics, leading to distorted identity information in the generated videos.
Therefore, we propose a new learning framework to mitigate this issue.

\section{Method}

\subsection{Problem Formulation}

\textbf{Problem.} Let $I_{\text{ref}}$ and $\mathcal{P}$ denote a reference face image and a text prompt describing the semantics of the target video, respectively.
The goal of identity-preserving text-to-video (IPT2V) generation is to create a video $\mathcal{V}$ under the condition of $I_{\text{ref}}$ and $\mathcal{P}$.
Thus, $\mathcal{V}$ should satisfy: \textbf{i)} the semantic information of $\mathcal{V}$ is aligned with $\mathcal{P}$ (i.e., textual alignment); and \textbf{ii)} most importantly, the facial identity information of the subject in $\mathcal{V}$ is consistent with $I_{\text{ref}}$.
The generation process can be formalized as:
\begin{equation}
\mathcal{V}=\mathcal{G}\left(\mathbf{Z}\sim\mathcal{N}(\mu,\sigma^{2}),\phi(I_{\text{ref}}),\mathcal{P}\right),
\end{equation}
where $\mathcal{G}$ is a text-to-video foundational generative model (e.g., Wan~\cite{Wan}).
$\mathbf{Z}$ is a prior state sampled from the Gaussian prior distribution.
$\phi$ denotes a function used to encode the identity information of $I_{\text{ref}}$.
For the above equation, the foundational model $\mathcal{G}$ determines the degree of semantic alignment between $\mathcal{V}$ and $\mathcal{P}$.
Therefore, researchers only need to select the strongest pretrained model and keep its original prior knowledge (e.g., LoRA Adapter~\cite{LoRA}) during training, which is not the focus of the IPT2V task.
For the function $\phi$, which determines the fidelity of facial identity information, i.e., the consistency of facial structure and the fidelity of facial texture details in the generated video $\mathcal{V}$.
Thus, this is a critical issue in the IPT2V task, and researchers are dedicated to constructing a robust $\phi$ that accurately represents the subject's identity information.

Recent state-of-the-art works have made various attempts and proposed diverse $\phi$ to improve the performance of IPT2V.
For example, ConsisID~\cite{ConsisID} proposed a global facial extractor and a local facial extractor to extract low-frequency structures and high-frequency details of the reference image $I_{\text{ref}}$, respectively.
Magic Mirror~\cite{Magicmirror} designed a dual-branch facial feature extractor to capture both identity and structural features.
However, they may struggle to handle situations involving complex facial dynamics, such as drastic changes in facial poses and emotions, or facial occlusions, resulting in distorted facial identity and facial structure in the generated videos.
The reason behind this phenomenon is that the encoded identity information can only represent a single pose view of the input image, failing to capture global pose information.

\textbf{Main Idea.} The identity information encoder could be partitioned into two parts: a basic facial identity encoder $\phi_{\text{bas}}$ and a global facial pose encoder $\phi_{\text{gfp}}$.
The former aims to encode the single-view facial structure information and facial texture details as existing methods do, and the latter aims to capture global facial pose representation.
Formally, our generation process is defined as:
\begin{equation}\label{mian-idea}
\mathcal{V}=\mathcal{G}\left(\mathbf{Z}\sim\mathcal{N}(\mu,\sigma^{2}),[\phi_{\text{bas}}(I_{\text{ref}}),\phi_{\text{gfp}}(I_{\text{ref}})],\mathcal{P}\right).
\end{equation}
Accordingly, there are two questions that need to be solved:
\begin{itemize}
    \item[-] Global facial pose encoder $\phi_{\text{gfp}}$. Representing faithful global facial pose from the input single-view reference image $I_{\text{ref}}$ as introduced in Sec.~\ref {gfprl}.
    \item[-] Automatic facial video dataset pipeline $P_{f}$. Collecting and preprocessing the video data with large changes in facial poses for training $\phi_{\text{gfp}}$ as introduced in Sec.~\ref {pipeline}.
\end{itemize}

 \begin{figure*}[t]
\centering{\includegraphics[width=\linewidth]{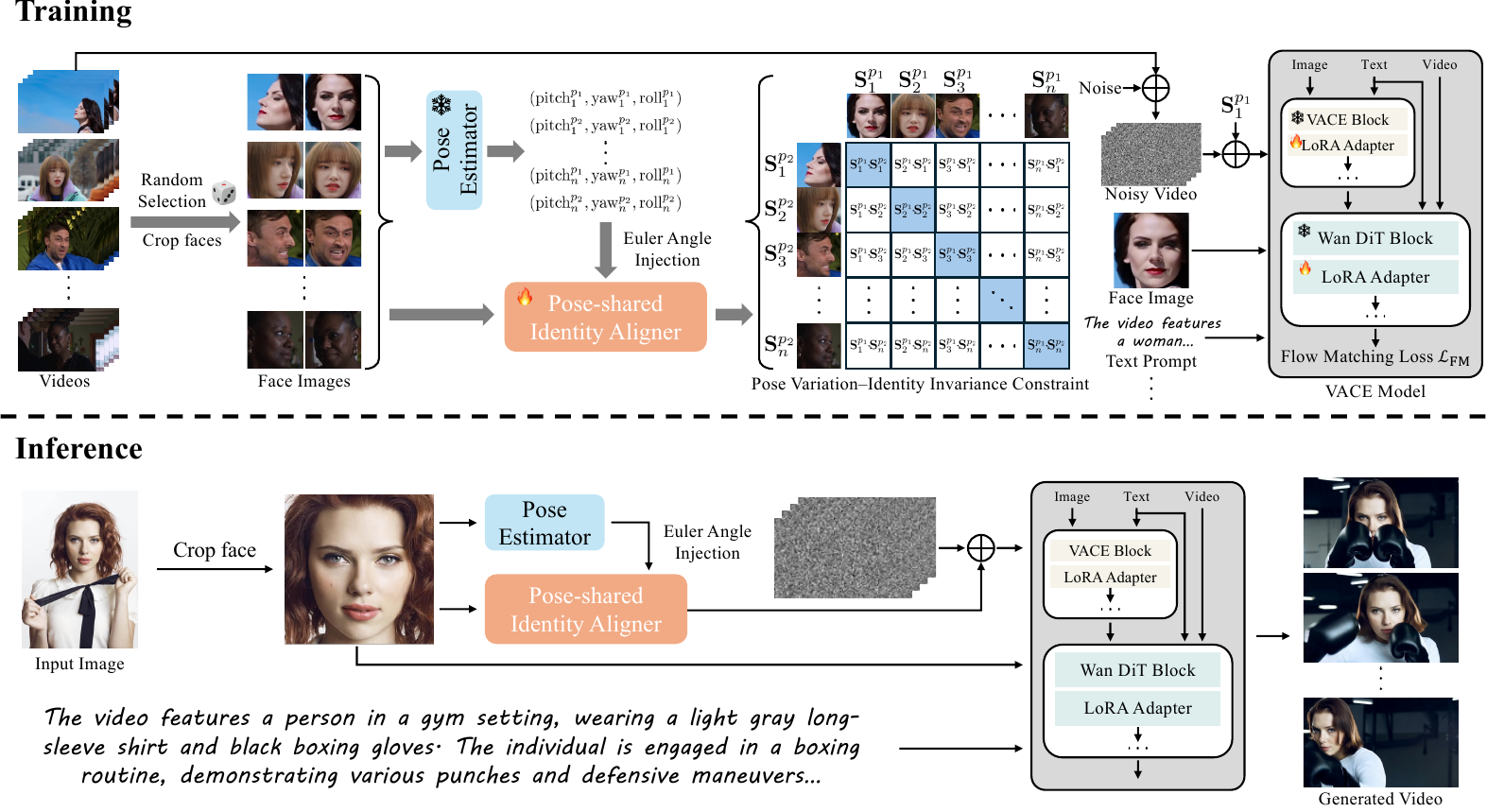}}
	\caption{The framework of FaithfulFaces. During training, given $n$ input videos per iteration, FaithfulFaces first randomly samples and crops two face images from each video. The cropped face images are then fed into a pose estimator to regress the three Euler angles, e.g., $(\text{pitch}_1^{p_1},\text{yaw}_1^{p_1},\text{roll}_1^{p_1})$, where ${p_1}$ and ${p_2}$ are simply used to mark two different poses. Next, the predicted Euler angles and the face images are then jointly fed into a pose-shared identity aligner, yielding $2n$ refined facial representations (e.g., $\mathbf{S}_1^{p_1}$), which are utilized to form a pose variation–identity invariance constraint. 
    Finally, the refined representations are injected into the patch-embedded video tokens as identity conditioning for joint optimization. During inference, FaithfulFaces encodes a global facial pose feature from a single face image and incorporates it into the generative model to produce videos.
	}
	\label{fig:2}
\end{figure*}

\subsection{Overview Framework}\label{Overview}
The overview framework of FaithfulFaces is illustrated in Fig~.\ref{fig:2}, which is divided into the training stage and the inference stage.
For the training stage, assuming there are $n$ videos as input for each training iteration, we first randomly sample and crop two face images from each video.
Subsequently, the cropped face images are fed into a pose estimator to regress the three Euler angles (i.e., pitch, yaw, roll) of the facial pose for each face image. 
These Euler angles, along with the face images, are then fed into our proposed pose-shared identity aligner to output $2n$ refined facial representations.
Furthermore, the $2n$ facial representations from all video samples can be combined into two batches of facial data to form a pose variation–identity invariance constraint. 
In this constraint, face images from the same identity with different poses are paired as positive samples (diagonal pairs), while those of different identities are paired as negative samples.
Finally, the output global facial pose representation is incorporated into the video generation backbone as identity conditioning. Specifically, the 5,120-dimensional identity representation is reshaped into a single token and zero-padded along the sequence dimension to match the patch-embedded video tokens. The resulting identity conditioning is added to the video token sequence before the DiT Transformer blocks, rather than directly to the raw noisy latent.
In practice, we utilize the VACE~\cite{VACE} as our foundational model and employ a LoRA training mode to fit these new data, where the VACE blocks are the basic facial identity encoder $\phi_{\text{bas}}$ in Eq.~(\ref{mian-idea}) to extract the single-view facial structure information and facial texture details.

During inference, users need only supply a single face image. The pose estimator regresses the Euler angles from this image, and both the angles and the image are passed to a well-trained identity aligner to generate the global facial pose representation.
The resulting representation is then injected into the patch-embedded video tokens as identity conditioning, guiding the generative model to produce identity-consistent videos conditioned on the reference image and text prompt.

\begin{figure*}[t]
\centering{\includegraphics[width=\linewidth]{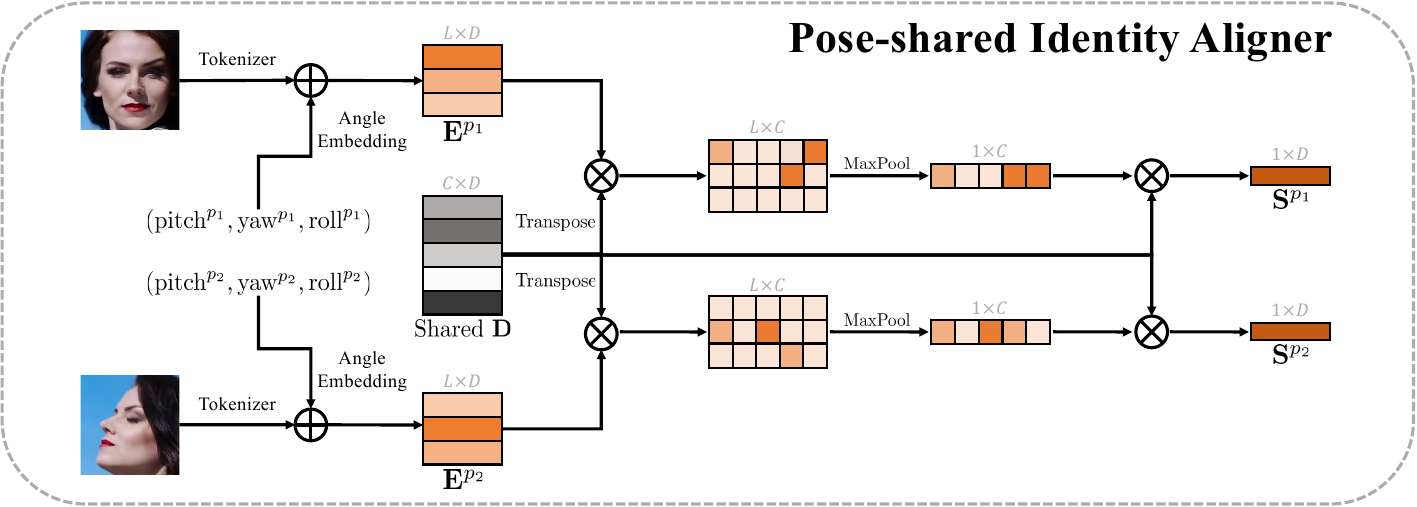}}
	\caption{Architecture of the pose-shared identity aligner. Initially, the input face images are tokenized into sequential face embeddings. The corresponding Euler angles are encoded as Euler angle embeddings and injected into the face embeddings, resulting in the combined embeddings $\mathbf{E}^{p_1}$ and $\mathbf{E}^{p_2}$.
    Then, a pose-shared dictionary $\mathbf{D}$ is employed to refine and align $\mathbf{E}^{p_1}$ and $\mathbf{E}^{p_2}$, yielding the global facial pose representations $\mathbf{S}^{p_1}$ and $\mathbf{S}^{p_2}$. Finally, these representations serve as pose-faithful facial priors for the foundational generative model.
	}
	\label{fig:3}
\end{figure*}

\subsection{Pose-shared Identity Aligner for Global Facial Pose Representation}\label{gfprl}
For the above framework, the most critical question is how to design and train the pose-shared identity aligner, i.e., encoder $\phi_{\text{gfp}}$ in Eq.~(\ref{mian-idea}), to represent robust global facial pose information.

Inspired by dictionary learning~\cite{dir-1,dir-2}, the key of our pose-shared identity aligner is to align the different facial poses into a refined dictionary space.
Fig.~\ref{fig:3} shows the architecture of the pose-shared identity aligner, which can receive face images with various poses and tokenize them into sequential face embeddings.
These vanilla face embeddings contain only implicit pixel-level facial pose information, which hinders the model's ability to perceive facial pose.
Thus, we aim to provide explicit pose information to guide the model's representation.
Specifically, we utilize a pretrained facial pose estimator (6DRepNet~\cite{6DRepNet} in practice) to regress three Euler angles: pitch, yaw, and roll.
Notably, Euler angles possess a periodic property, which makes it natural to generate their embeddings using the timestep encoding method employed in diffusion models~\cite{ddpm}.
As shown in Fig.~\ref{fig:3}, we inject the Euler angle embeddings into the vanilla face embeddings to generate two new embeddings, marked as $\mathbf{E}^{p_1}\in \mathbb{R}^{L\times D}$ and $\mathbf{E}^{p_2}\in \mathbb{R}^{L\times D}$, where $L$ and $D$ denote the sequence length and dimensionality.
${p_1}$ and ${p_2}$ are simply used to mark two different poses.
With these embeddings, we then define a learnable pose-shared dictionary matrix $\mathbf{D}\in \mathbb{R}^{C\times D}$, where $C$ indicates the number of dictionary elements.
Subsequently, $\mathbf{E}^{p_1}\in \mathbb{R}^{L\times D}$ and $\mathbf{E}^{p_2}\in \mathbb{R}^{L\times D}$ are projected into a dictionary space by calculating the correlation between each face embedding and $\mathbf{D}$ to obtain the correlation matrices, which can be further capsuled into two dictionary weights $\mathbf{W}^{p_1}$ and $\mathbf{W}^{p_2}$:
\begin{equation}
    \mathbf{W}^{p_1} = \text{MaxPool}(\mathbf{E}^{p_1}\otimes \mathbf{D}^{\top})\in \mathbb{R}^{1\times C},
    \mathbf{W}^{p_2} = \text{MaxPool}(\mathbf{E}^{p_2}\otimes \mathbf{D}^{\top})\in \mathbb{R}^{1\times C},
\end{equation}
where $\text{MaxPool}(\cdot)$ denotes a max pooling operation empirically determined in Appendix~\ref{abl_pooling}.
$\otimes$ means matrix multiplication.
Finally, these dictionary weights can be used to obtain the global facial pose representations $\mathbf{S}^{p_1}$ and $\mathbf{S}^{p_2}$:
\begin{equation}
    \mathbf{S}^{p_1} \!=\! (\mathbf{W}^{p_1}\otimes \mathbf{D})\!\in\! \mathbb{R}^{1\times D},\mathbf{S}^{p_2} \!=\! (\mathbf{W}^{p_2}\otimes \mathbf{D})\!\in\! \mathbb{R}^{1\times D}.
\end{equation}
To optimize this aligner, we observe that the two batches of input facial data with different poses can exactly form a CLIP-like contrastive paradigm, as shown in the upper part of Fig.~\ref{fig:2}.
Thus, we apply the most commonly used contrastive learning~\cite{clip} to train our aligner:
\begin{equation}
    \mathcal{L}_{\text{PIA}} = -\frac{1}{n}\!\!\sum_{i=1}^{n} \log \frac{\exp(\text{sim}(\mathbf{S}_i^{p_1},\mathbf{S}_i^{p_2})/\tau)}{\sum_{j=1}^n\exp(\text{sim}(\mathbf{S}_i^{p_1},\mathbf{S}_j^{p_2})/\tau)} 
    - \frac{1}{n}\!\!\sum_{i=1}^{n} \log \frac{\exp(\text{sim}(\mathbf{S}_i^{p_2},\mathbf{S}_i^{p_1})/\tau)}{\sum_{j=1}^n\exp(\text{sim}(\mathbf{S}_i^{p_2},\mathbf{S}_j^{p_1})/\tau)},
\end{equation}
where $n$ is the number of matched identity pairs in each training mini-batch, $\text{sim}(\cdot,\cdot)$ denotes the cosine similarity function, and $\tau$ is a learnable temperature parameter with the default setting of \cite{clip}.
During the whole training process, we integrate $\mathcal{L}_{\text{PIA}}$ with the objective of the generative model (i.e., flow matching~\cite{rf}) $\mathcal{L}_{\text{FM}}$ to reach the full optimization objective: 
\begin{equation}
\mathcal{L}_{\text{total}}=\mathcal{L}_{\text{PIA}}+\mathcal{L}_{\text{FM}}.
\end{equation}
In practice, $\mathcal{L}_{\text{PIA}}$ and $\mathcal{L}_{\text{FM}}$ are responsible for their respective tasks during the training process. $\mathcal{L}_{\text{PIA}}$ is dedicated to constraining the alignment of different poses, while $\mathcal{L}_{\text{FM}}$ is dedicated to constraining the LoRA parameters to adapt to the input's global facial pose representation. 
This approach ensures that the different loss functions can focus on handling their specific tasks.

\begin{rmk} \textnormal{(\textbf{Deep insights and observations})}
\textit{Our design of the pose-shared identity aligner is not only intuitive but also admits a theoretical justification. Recall that $\mathcal{L}_{\textnormal{PIA}}$ is equivalent to the InfoNCE loss~\cite{cpc}, which provides a lower bound of the mutual information:
\begin{equation}
    I(\mathbf{S}^{p_1}; \mathbf{S}^{p_2}) \geq \log(n) - \mathcal{L}_{\textnormal{PIA}}.
\end{equation}
This inequality implies that minimizing $\mathcal{L}_{\textnormal{PIA}}$ is not only aligning pose-variant embeddings but also maximizing the shared identity information across different poses. 
Hence, our aligner has an information-theoretic guarantee: the learned global representation cannot collapse unless $I(\mathbf{S}^{p_1}; \mathbf{S}^{p_2})$ vanishes.
From the experimental observation, the visualization of the encoded facial identity in Fig.~\ref{fig:vis-tsne} confirms the above insights.
Furthermore, the learned dictionary reveals meaningful activation patterns, wherein images with similar poses tend to frequently activate particular dictionary elements, as illustrated in Fig.~\ref{fig:dir_stat}.
This indicates that the learned dictionary facilitates robust representation of faces across a wide range of poses.
}
\end{rmk}

\begin{figure*}[t]
\centering{\includegraphics[width=\linewidth]{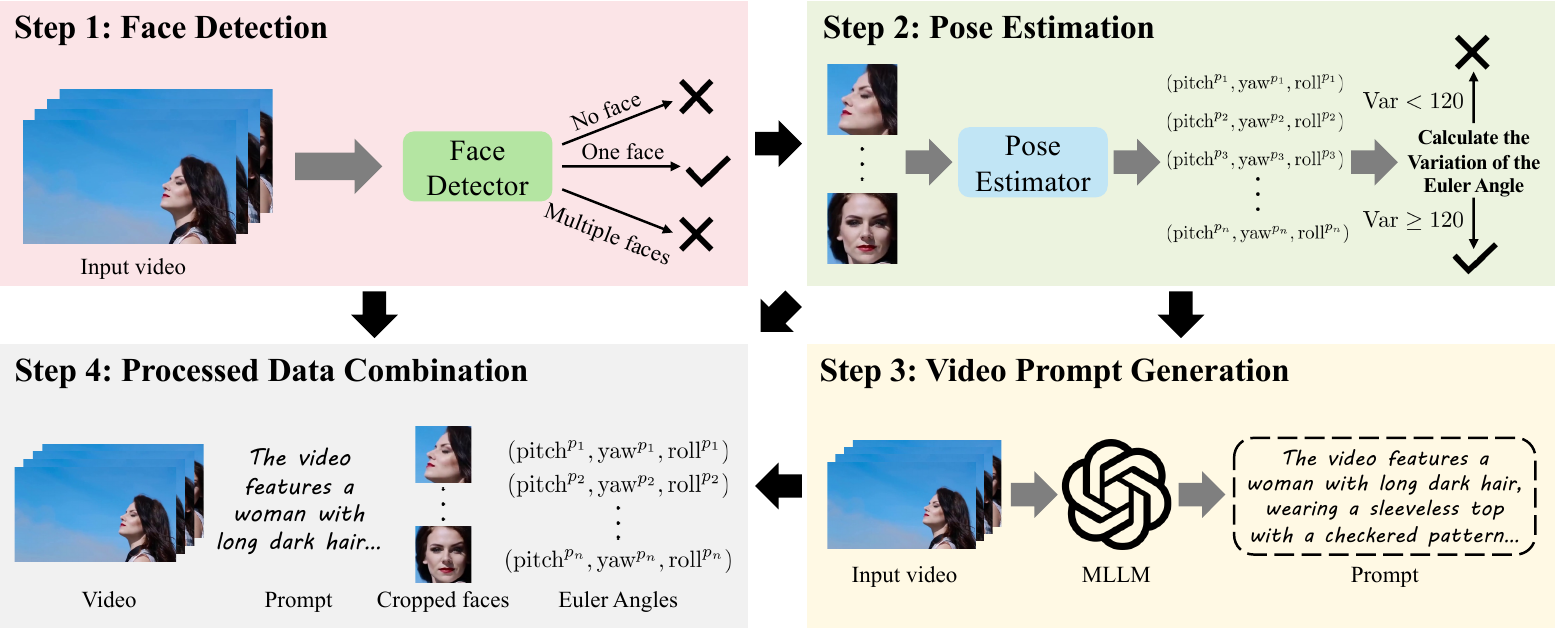}}
	\caption{Dataset collection and processing pipeline. In Step 1, videos without face or with multiple faces are filtered out. Step 2 aims to select videos that exhibit significant variations in facial pose.
    For Step 3, we generate a text prompt for each selected video using MLLM. Step 4  ultimately integrates these fragmented data into a cohesive whole.
	}
	\label{fig:dataset}
\end{figure*}

\subsection{Dataset Construction}\label{pipeline}
Beyond framework design, a critical challenge persists: constructing a video dataset with significant variations in facial poses for training our proposed pose-shared identity aligner.
This is because ordinary facial micro-movements or static videos are insufficient to satisfy our training requirements.

To address this issue, we construct a new dataset collection and processing pipeline.
Note that this part omits standard data collection and preprocessing procedures that have been widely adopted in previous works~\cite{VACE,Hunyuancustom,Phantom}, such as video clip segmentation, resolution standardization, OCR filter, aesthetic filter, clarity filter, etc.
The original videos are sourced from OpenHumanVid~\cite{Openhumanvid}, a publicly documented human-centric video dataset obtained through its official application-and-review process. We further curate and process these videos to construct our task-oriented training dataset, without independently collecting additional human-subject videos. The resolution of each video is standardized to $832 \times 480$ pixels.
Fig.~\ref{fig:dataset}  illustrates our dataset collection and processing pipeline, which consists of four steps: face detection, pose estimation, video prompt generation, and processed data combination.

\textbf{Face Detection.}
Since our work only focuses on the single-subject video generation task, we first need to filter out two types of videos: videos without face and videos with multiple faces.
Specifically, we utilize InsightFace~\cite{InsightFace} for face detection on each video frame. Videos are filtered out if more than one face is detected in any single frame. Additionally, videos in which no faces are detected throughout the entire sequence are also excluded.

\textbf{Pose Estimation.} 
This step constitutes the core of the entire dataset pipeline, aiming to select videos that exhibit significant variations in facial pose.
Taking a video $\mathcal{V}$ as an example, we first use the facial bounding boxes obtained in the previous step to crop the face regions from each video frame. 
These cropped face regions are then fed into the pose estimator 6DRepNet to predict three Euler angles for each detected face.
Note that in practice, we enlarge the bounding boxes by a factor of 1.5 to predict Euler angles more accurately.
Next, the three Euler angles for each face are stored separately in three lists, denoted as $\mathcal{X}_{\text{pitch}}$, $\mathcal{X}_{\text{yaw}}$, and $\mathcal{X}_{\text{roll}}$, and we can calculate the variation of Euler angles throughout the entire video:
\begin{equation}
    \text{Var} = [\max(\mathcal{X}_{\text{pitch}})-\min(\mathcal{X}_{\text{pitch}})] \\
    + [\max(\mathcal{X}_{\text{yaw}})-\min(\mathcal{X}_{\text{yaw}})] 
    + [\max(\mathcal{X}_{\text{roll}})-\min(\mathcal{X}_{\text{roll}})],
\end{equation}
where $\max(\cdot)$ and $\min(\cdot)$ represent the maximum and minimum values in the list, respectively.
Furthermore, it is necessary to determine a reliable variation threshold to filter out qualified videos.
To determine this threshold, we first randomly sample 2000 videos from the output of step 1 and manually annotate them.
Our criterion for qualified videos is that the facial pose in the video must show at least a transition from frontal to profile (or vice versa), or exhibit significant up-and-down movement. Videos meeting these criteria are labeled as qualified, and we finally determine that the threshold is 120.
With this threshold, we can retain videos with large facial pose changes; that is, $\text{Var}>120$ is qualified, while $\text{Var}<120$ is discarded.

\textbf{Video Prompt Generation.}
After collecting qualified videos, we need to generate a text prompt for each video.
Here, we use Qwen2.5-VL~\cite{Qwen2.5-VL} to generate information-rich text prompts for qualified videos, focusing on describing the subjects’ appearance, actions, and background.
We then perform extensive manual calibration and refinement to improve the accuracy of text prompts.

\textbf{Processed Data Combination.}
After the above three steps of data screening and preprocessing, we ultimately integrate these fragmented data into a cohesive whole.
As shown in step 4 of Fig.~\ref{fig:dataset}, each sample in our well-constructed dataset contains four elements: video, text prompt, cropped face images, and Euler angles.
We manually check all processed data to ensure that all videos are qualified enough, and ultimately generate 51,624 samples for model training.

\begin{table*}[t]
	\centering
	\caption{Quantitative comparison with evaluated baselines.}\label{tab:1}
	\scalebox{0.95}{
         \begin{tabular}{c|c|c|c|c|c}
        \hline
        Methods & Open-Source & FaceSim-Cur~$\uparrow$ &  FaceSim-Arc~$\uparrow$ & FID~$\downarrow$ & CLIPScore~$\uparrow$ \\
        \hline
        \hline
        Vidu~\cite{Vidu} & $\times$  &  0.293 & 0.278 & 234.65 & 30.08 \\
        Kling~\cite{Kling} & $\times$  & 0.447 & 0.416 & 194.80 & 33.06 \\
        ConsisID~\cite{ConsisID} & $\checkmark$  &  0.365 & 0.350 & 205.03 & 30.29 \\
        VACE-14B~\cite{VACE} & $\checkmark$ & 0.403  & 0.382 & 191.02 & 31.83 \\
        HunyuanCustom~\cite{Hunyuancustom} & $\checkmark$ & 0.453 & 0.432 & 187.32 & 31.36 \\
        Phantom-14B~\cite{Phantom} & $\checkmark$ & 0.484 & 0.456 & 214.99 & 29.67 \\
        Concat-ID-Wan~\cite{Concat-ID} & $\checkmark$ & 0.408 & 0.387 & 189.55 & 31.49 \\
        SkyReels-A2~\cite{Skyreels-a2} & $\checkmark$ & 0.410 & 0.384 & 237.29 & 28.10 \\
        Stand-In~\cite{Stand-In} & $\checkmark$  & 0.415 & 0.395 & 196.21 & 30.38 \\
        MAGREF~\cite{MAGREF} & $\checkmark$ & 0.417 & 0.392 & 207.69 & 31.13 \\
        \textbf{FaithfulFaces (Ours)} & $\checkmark$ & \textbf{0.568} & \textbf{0.542} & \textbf{164.24} & \textbf{33.93}\\
			\hline
	\end{tabular}}
\end{table*}

\begin{figure*}[t]
\centering{\includegraphics[width=\linewidth]{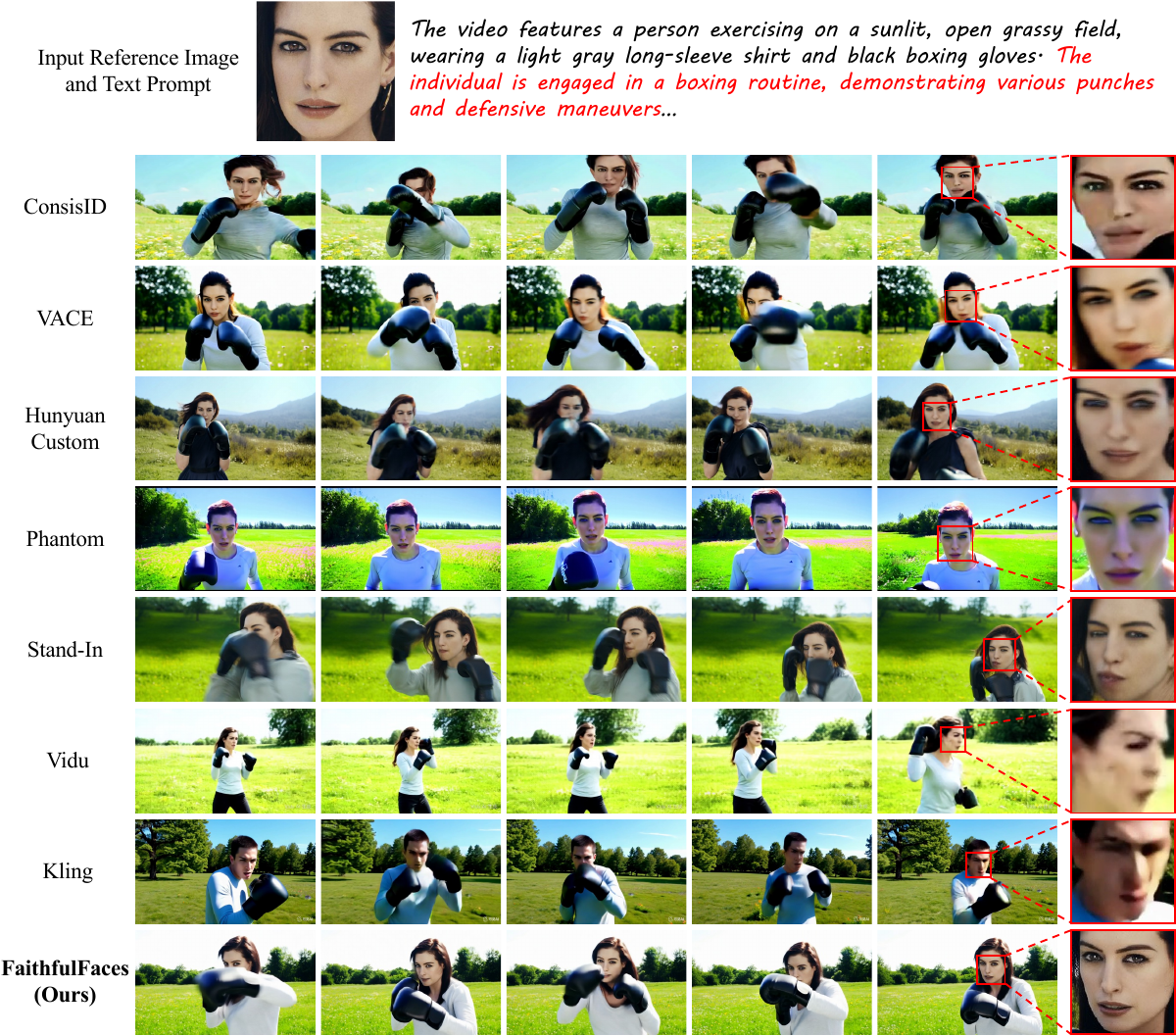}}
	\caption{Visual comparisons of different methods. The goal is to generate a video of a person engaging in a boxing routine, characterized by diverse facial poses and instances of facial occlusion. In contrast to these state-of-the-art methods, FaithfulFaces produces a video of superior quality, exhibiting clear facial structures and consistent identity preservation.
	}
	\label{fig:vis_1}
\end{figure*}

\section{Experiments}

\subsection{Implementation Details}\label{Implementation}
Our FaithfulFaces framework utilizes the DiT-based generative model VACE-14B~\cite{VACE} as our foundational model.
For the pose-shared identity aligner, the number of dictionary elements of $\mathbf{D}$ is set to 4096, empirically determined in Appendix~\ref{dic_num}.
We set the resolution of each video to $832\times 480$ pixels and extract 81 consecutive frames for training.
In the training phase, we use the LoRA training mode with rank 128 to fit new data.
The whole framework is trained on 32 NVIDIA H20 GPUs with a batch size of 32.
In addition, we set an independent batch size of 1024 for the pose-shared identity aligner to perform adequate pose alignment, and the total number of training steps is set to 5000.

\textbf{Evaluation details.} 
We conduct experiments and evaluations on several face images used in ConsisID~\cite{ConsisID}, which consist of 30 persons, and we randomly sample one image for each identity.
We then construct 20 challenging text prompts that drive the models to generate videos featuring significant facial pose variations, expression changes, and facial occlusions across diverse scenarios. The details can be found in Appendix~\ref{Prompt}.
We consider four standard evaluation metrics that are used in prior works~\cite{Moviegen,ConsisID} to measure the quality of generated videos.
\textbf{FaceSim-Arc} and \textbf{FaceSim-Cur} are employed to assess identity preservation by measuring feature discrepancies between face regions in the generated videos and those in real face images within the ArcFace~\cite{Arc} and CurricularFace~\cite{Cur} feature spaces.
For visual quality, we utilize the commonly used \textbf{FID}~\cite{fid} by calculating feature differences in the face regions between the generated frames and real face images within the InceptionV3~\cite{InceptionV3} feature space.
For textual alignment, we utilize the \textbf{CLIPScore}~\cite{CLIPScore} to measure the similarity between videos and text prompts.

\subsection{Baseline Comparisons}
We compare our FaithfulFaces with the current state-of-the-art methods, including two commercial products (Vidu~\cite{Vidu}, Kling~\cite{Kling}) and eight open-source models (ConsisID ~\cite{ConsisID}, VACE~\cite{VACE}, HunyuanCustom~\cite{Hunyuancustom}, Phantom~\cite{Phantom}, Concat-ID-Wan~\cite{Concat-ID}, SkyReels-A2~\cite{Skyreels-a2}, Stand-In~\cite{Stand-In}, MAGREF~\cite{MAGREF}).
For these open-source methods, ConsisID is based on CogVideoX-5B~\cite{CogVideoX}, HunyuanCustom is based on HunyuanVideo~\cite{Hunyuanvideo}, VACE, Phantom, Concat-ID-Wan, SkyReels-A2, Stand-In, and MAGREF are based on Wan~\cite{Wan}, providing diversity for evaluation and comparison.
For each method, we generate 600 videos (30 persons $\times$ 20 prompts) for evaluation, which is larger than the current top-tier community standard (e.g., MAGREF~\cite{MAGREF} generates 120 videos for evaluation).

\textbf{Quantitative results.}
Tab.~\ref{tab:1} lists the quantitative results of different methods. From these results, we can observe that FaithfulFaces achieves the best IPT2V performance under four evaluation metrics.
In particular, FaithfulFaces gains considerable performance improvements in the FaceSim-Cur and FaceSim-Arc metrics used to measure identity preservation of generated videos.
This improvement can be attributed to FaithfulFaces’s ability to provide a robust global facial pose prior for foundational generative models, enabling more effective identity preservation.

\textbf{Qualitative results.}
Fig.~\ref{fig:vis_1} provides some visual comparisons of our FaithfulFaces against seven baselines, including a case of generating a video of a person engaging in a boxing routine.
We can first observe that the subjects in the videos generated by different methods have various facial pose changes and even facial occlusion due to intense movements.
Furthermore, we discover that these open-source and commercial methods exhibit varying degrees of identity distortion and facial collapse as the subject moves.
In contrast, our FaithfulFaces can output a high-quality video with clear facial structures and consistent identity details.
Similar observations are made in Appendix~\ref{app_vis}, where more visual results are provided. 
Video demos are included in the Supplementary Material.

\subsection{Ablation Studies}
We evaluate the effects of the key components in FaithfulFaces, including the pose-shared identity aligner (Aligner) and the injection of Euler angle embeddings (Euler).
For a fair comparison, all ablation variants are fine-tuned from the same VACE-14B backbone using identical 51,624 training samples, LoRA configurations, optimization settings, and training steps. In particular, the ``w/o (Aligner, Euler)'' variant undergoes the same fine-tuning procedure as FaithfulFaces, thereby isolating the improvements introduced by the proposed components from those arising from additional training data.
The results are presented in Tab.~\ref{tab:Ablation}, from which we draw the following conclusions:
\textbf{i)} Identity Aligner is effective and yields substantial performance improvements, as it represents global facial pose information from the input single-view reference image, thereby enhancing the identity consistency of the generated videos.
\textbf{ii)} The inclusion of Euler Embedding yields further improvements, confirming the feasibility and effectiveness of explicitly injecting pose information.
More ablation studies are provided in Appendix~\ref{dic_num}, \ref{vis-abl}, \ref{abl_pooling}, and \ref{identity-features}, including qualitative analysis, dictionary elements, pooling operation type, and identity features.
Additionally, in Appendix~\ref{non-frontal-discuss} and~\ref{aligner-euler-discuss}, we investigate the robustness to non-frontal views and the identity aligner's sensitivity to Euler angle variations.

\begin{table}[h]
	\centering
	\caption{Ablation study of the key components in FaithfulFaces.}\label{tab:Ablation}
	\scalebox{1.0}{
         \begin{tabular}{cc|c|c|c|c}
        \hline
        Aligner & Euler & FaceSim-Cur~$\uparrow$ &  FaceSim-Arc~$\uparrow$ & FID~$\downarrow$ & CLIPScore~$\uparrow$ \\
        \hline
        \hline
    $\checkmark$ & $\checkmark$ & \textbf{0.568} & \textbf{0.542} & \textbf{164.24} & \textbf{33.93} \\
	$\checkmark$ & $\times$ & 0.522 & 0.497 & 173.57 & 33.31 \\
        $\times$ & $\times$ & 0.437 & 0.414 & 186.82 & 32.05 \\
        \hline
	\end{tabular}}
\end{table}

\begin{figure*}[h]
\centering{\includegraphics[width=\linewidth]{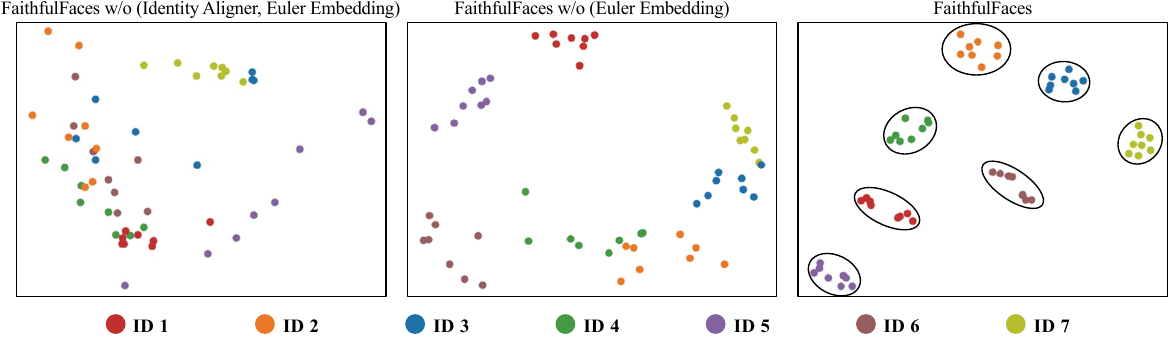}}
	\caption{Visualization of the encoded facial identity (ID). FaithfulFaces demonstrates promising ID separability, indicating that its encoded identity representation exhibits high faithfulness and fidelity.
	}
	\label{fig:vis-tsne}
\end{figure*}

\textbf{Visualization of the Encoded Facial Identity.}
Fig.~\ref{fig:vis-tsne} visualizes the distribution of encoded facial identities under different settings.
Specifically, we randomly select 7 videos with different identities that are not included in the training data, and sample 8 face images with different facial poses from each video.
All sampled face images are then encoded by the different methods, and the encoded features are projected into a 2D space by t-SNE~\cite{tsne}.
We can observe that \textit{FaithfulFaces w/o (Identity Aligner, Euler Embedding)} suffer from the collapse of facial identity due to the absence of global facial pose awareness.
At the same time, \textit{FaithfulFaces w/o (Euler Embedding)} alleviates the collapse of facial identity, but the discriminability of its facial identity representation remains limited.
In contrast, \textit{FaithfulFaces} shows promising identity separability, demonstrating its faithful facial identity and naturally enhancing the performance of IPT2V.

\section{Conclusion}
In this paper, we have proposed FaithfulFaces, a pose-faithful facial identity preservation learning framework for IPT2V.
FaithfulFaces is motivated by the observation that existing methods often struggle to handle some intricate facial dynamic scenarios, largely due to their insufficient awareness of global facial pose.
To encode the global facial pose representation from the input single-view face image, we propose a pose-shared identity aligner that refines and aligns distinct facial poses by a pose-shared dictionary and a pose variation–identity invariance constraint with Euler angle embedding learning.
In particular, we construct a task-oriented, high-quality dataset with substantial facial pose diversity for robust training.
Extensive experiments validate the effectiveness of FaithfulFaces.

\noindent\textbf{Limitations.}
Despite its effectiveness, FaithfulFaces still has several limitations. First, our framework primarily focuses on single-subject IPT2V generation, and its applicability to multi-subject scenarios remains unexplored. Second, although the proposed pose-shared identity aligner improves identity preservation under substantial pose variations, extreme head rotations and severe facial occlusions may still lead to identity degradation. Finally, since our framework builds upon a pretrained video generation backbone, certain visual artifacts and motion inconsistencies inherited from the backbone may persist. Addressing these limitations, particularly multi-subject identity preservation and robustness under challenging facial dynamics, will be explored in future work.

\section*{Acknowledgements}
This work was supported by the 2025 Tencent Rhino-bird Research Elite Program, the National Natural Science Foundation of China under Grant 62476133, and the Fundamental Research Funds for the Central Universities under Grant 11300-312200502507.

{\small
\bibliographystyle{ieee_fullname}
\bibliography{example_paper}

@inproceedings{CogVideoX,
  title={CogVideoX: Text-to-Video Diffusion Models with An Expert Transformer},
  author={Yang, Zhuoyi and Teng, Jiayan and Zheng, Wendi and Ding, Ming and Huang, Shiyu and Xu, Jiazheng and Yang, Yuanming and Hong, Wenyi and Zhang, Xiaohan and Feng, Guanyu and others},
  booktitle={The Thirteenth International Conference on Learning Representations},
  year={2025}
}

@article{Wan,
      title={Wan: Open and Advanced Large-Scale Video Generative Models}, 
      author={Team Wan and Ang Wang and Baole Ai and Bin Wen and Chaojie Mao and Chen-Wei Xie and Di Chen and Feiwu Yu and Haiming Zhao and Jianxiao Yang and Jianyuan Zeng and Jiayu Wang and Jingfeng Zhang and Jingren Zhou and Jinkai Wang and Jixuan Chen and Kai Zhu and Kang Zhao and Keyu Yan and Lianghua Huang and Mengyang Feng and Ningyi Zhang and Pandeng Li and Pingyu Wu and Ruihang Chu and Ruili Feng and Shiwei Zhang and Siyang Sun and Tao Fang and Tianxing Wang and Tianyi Gui and Tingyu Weng and Tong Shen and Wei Lin and Wei Wang and Wei Wang and Wenmeng Zhou and Wente Wang and Wenting Shen and Wenyuan Yu and Xianzhong Shi and Xiaoming Huang and Xin Xu and Yan Kou and Yangyu Lv and Yifei Li and Yijing Liu and Yiming Wang and Yingya Zhang and Yitong Huang and Yong Li and You Wu and Yu Liu and Yulin Pan and Yun Zheng and Yuntao Hong and Yupeng Shi and Yutong Feng and Zeyinzi Jiang and Zhen Han and Zhi-Fan Wu and Ziyu Liu},
      journal = {arXiv preprint arXiv:2503.20314},
      year={2025}
}

@inproceedings{LoRA,
  title={LoRA: Low-Rank Adaptation of Large Language Models},
  author={Hu, Edward J and Wallis, Phillip and Allen-Zhu, Zeyuan and Li, Yuanzhi and Wang, Shean and Wang, Lu and Chen, Weizhu and others},
  booktitle={International Conference on Learning Representations},
  year = {2022}
}

@inproceedings{ConsisID,
  title={Identity-preserving text-to-video generation by frequency decomposition},
  author={Yuan, Shenghai and Huang, Jinfa and He, Xianyi and Ge, Yunyang and Shi, Yujun and Chen, Liuhan and Luo, Jiebo and Yuan, Li},
  booktitle={Proceedings of the Computer Vision and Pattern Recognition Conference},
  pages={12978--12988},
  year={2025}
}

@article{Fantasyid,
  title={Fantasyid: Face knowledge enhanced id-preserving video generation},
  author={Zhang, Yunpeng and Wang, Qiang and Jiang, Fan and Fan, Yaqi and Xu, Mu and Qi, Yonggang},
  journal={arXiv preprint arXiv:2502.13995},
  year={2025}
}

@inproceedings{clip,
  title={Learning transferable visual models from natural language supervision},
  author={Radford, Alec and Kim, Jong Wook and Hallacy, Chris and Ramesh, Aditya and Goh, Gabriel and Agarwal, Sandhini and Sastry, Girish and Askell, Amanda and Mishkin, Pamela and Clark, Jack and others},
  booktitle={International conference on machine learning},
  pages={8748--8763},
  year={2021},
  organization={PmLR}
}

@article{dir-1,
  title={Neural discrete representation learning},
  author={Van Den Oord, Aaron and Vinyals, Oriol and others},
  journal={Advances in neural information processing systems},
  volume={30},
  year={2017}
}

@inproceedings{dir-2,
  title={Multi-modal alignment using representation codebook},
  author={Duan, Jiali and Chen, Liqun and Tran, Son and Yang, Jinyu and Xu, Yi and Zeng, Belinda and Chilimbi, Trishul},
  booktitle={Proceedings of the IEEE/CVF Conference on Computer Vision and Pattern Recognition},
  pages={15651--15660},
  year={2022}
}

@inproceedings{6DRepNet,
  title={6d rotation representation for unconstrained head pose estimation},
  author={Hempel, Thorsten and Abdelrahman, Ahmed A and Al-Hamadi, Ayoub},
  booktitle={2022 IEEE International Conference on Image Processing (ICIP)},
  pages={2496--2500},
  year={2022},
  organization={IEEE}
}

@article{ddpm,
  title={Denoising diffusion probabilistic models},
  author={Ho, Jonathan and Jain, Ajay and Abbeel, Pieter},
  journal={Advances in neural information processing systems},
  volume={33},
  pages={6840--6851},
  year={2020}
}

@article{VACE,
    title = {VACE: All-in-One Video Creation and Editing},
    author = {Jiang, Zeyinzi and Han, Zhen and Mao, Chaojie and Zhang, Jingfeng and Pan, Yulin and Liu, Yu},
    journal = {arXiv preprint arXiv:2503.07598},
    year = {2025}
}

@inproceedings{fm,
  title={Flow Matching for Generative Modeling},
  author={Lipman, Yaron and Chen, Ricky TQ and Ben-Hamu, Heli and Nickel, Maximilian and Le, Matthew},
  booktitle={The Eleventh International Conference on Learning Representations},
  year={2023}
}

@inproceedings{rf,
  title={Flow Straight and Fast: Learning to Generate and Transfer Data with Rectified Flow},
  author={Liu, Xingchao and Gong, Chengyue and others},
  booktitle={The Eleventh International Conference on Learning Representations},
  year = {2023}
}

@article{Hunyuancustom,
  title={Hunyuancustom: A multimodal-driven architecture for customized video generation},
  author={Hu, Teng and Yu, Zhentao and Zhou, Zhengguang and Liang, Sen and Zhou, Yuan and Lin, Qin and Lu, Qinglin},
  journal={arXiv preprint arXiv:2505.04512},
  year={2025}
}

@article{Phantom,
  title={Phantom: Subject-consistent video generation via cross-modal alignment},
  author={Liu, Lijie and Ma, Tianxiang and Li, Bingchuan and Chen, Zhuowei and Liu, Jiawei and Li, Gen and Zhou, Siyu and He, Qian and Wu, Xinglong},
  journal={arXiv preprint arXiv:2502.11079},
  year={2025}
}

@article{Qwen2.5-VL,
  title={Qwen2.5-VL Technical Report},
  author={Bai, Shuai and Chen, Keqin and Liu, Xuejing and Wang, Jialin and Ge, Wenbin and Song, Sibo and Dang, Kai and Wang, Peng and Wang, Shijie and Tang, Jun and Zhong, Humen and Zhu, Yuanzhi and Yang, Mingkun and Li, Zhaohai and Wan, Jianqiang and Wang, Pengfei and Ding, Wei and Fu, Zheren and Xu, Yiheng and Ye, Jiabo and Zhang, Xi and Xie, Tianbao and Cheng, Zesen and Zhang, Hang and Yang, Zhibo and Xu, Haiyang and Lin, Junyang},
  journal={arXiv preprint arXiv:2502.13923},
  year={2025}
}

@article{Hunyuanvideo,
  title={Hunyuanvideo: A systematic framework for large video generative models},
  author={Kong, Weijie and Tian, Qi and Zhang, Zijian and Min, Rox and Dai, Zuozhuo and Zhou, Jin and Xiong, Jiangfeng and Li, Xin and Wu, Bo and Zhang, Jianwei and others},
  journal={arXiv preprint arXiv:2412.03603},
  year={2024}
}

@article{Id-animator,
  title={Id-animator: Zero-shot identity-preserving human video generation},
  author={He, Xuanhua and Liu, Quande and Qian, Shengju and Wang, Xin and Hu, Tao and Cao, Ke and Yan, Keyu and Zhang, Jie},
  journal={arXiv preprint arXiv:2404.15275},
  year={2024}
}

@misc{Kling,
	author={Kling},
	year={2026},
	title={Kling},
	howpublished = {\url{https://klingai.com/}}
}

@misc{Vidu,
	author={Vidu},
	year={2026},
	title={Vidu},
	howpublished = {\url{https://www.vidu.com}}
}

@misc{InsightFace,
	author={InsightFace},
	year={2025},
	title={InsightFace},
	howpublished = {\url{https://github.com/deepinsight/insightface}}
}

@inproceedings{scoresde,
  title={Score-Based Generative Modeling through Stochastic Differential Equations},
  author={Song, Yang and Sohl-Dickstein, Jascha and Kingma, Diederik P and Kumar, Abhishek and Ermon, Stefano and Poole, Ben},
  booktitle={International Conference on Learning Representations},
  year={2021}
}

@article{Moviegen,
  title={Movie gen: A cast of media foundation models},
  author={Polyak, Adam and Zohar, Amit and Brown, Andrew and Tjandra, Andros and Sinha, Animesh and Lee, Ann and Vyas, Apoorv and Shi, Bowen and Ma, Chih-Yao and Chuang, Ching-Yao and others},
  journal={arXiv preprint arXiv:2410.13720},
  year={2024}
}

@article{Seedance,
  title={Seedance 1.0: Exploring the Boundaries of Video Generation Models},
  author={Gao, Yu and Guo, Haoyuan and Hoang, Tuyen and Huang, Weilin and Jiang, Lu and Kong, Fangyuan and Li, Huixia and Li, Jiashi and Li, Liang and Li, Xiaojie and others},
  journal={arXiv preprint arXiv:2506.09113},
  year={2025}
}

@inproceedings{Animatediff,
  title={AnimateDiff: Animate Your Personalized Text-to-Image Diffusion Models without Specific Tuning},
  author={Guo, Yuwei and Yang, Ceyuan and Rao, Anyi and Liang, Zhengyang and Wang, Yaohui and Qiao, Yu and Agrawala, Maneesh and Lin, Dahua and Dai, Bo},
  booktitle={International Conference on Learning Representations (ICLR)},
  year={2024}
}

@inproceedings{dit,
  title={Scalable diffusion models with transformers},
  author={Peebles, William and Xie, Saining},
  booktitle={Proceedings of the IEEE/CVF international conference on computer vision},
  pages={4195--4205},
  year={2023}
}

@misc{Pika,
	author={Pika},
	year={2025},
	title={Pika},
	howpublished = {\url{https://pika.art/}}
}

@inproceedings{Arc,
  title={Arcface: Additive angular margin loss for deep face recognition},
  author={Deng, Jiankang and Guo, Jia and Xue, Niannan and Zafeiriou, Stefanos},
  booktitle={Proceedings of the IEEE/CVF conference on computer vision and pattern recognition},
  pages={4690--4699},
  year={2019}
}

@inproceedings{Cur,
  title={Curricularface: adaptive curriculum learning loss for deep face recognition},
  author={Huang, Yuge and Wang, Yuhan and Tai, Ying and Liu, Xiaoming and Shen, Pengcheng and Li, Shaoxin and Li, Jilin and Huang, Feiyue},
  booktitle={proceedings of the IEEE/CVF conference on computer vision and pattern recognition},
  pages={5901--5910},
  year={2020}
}

@article{fid,
  title={Gans trained by a two time-scale update rule converge to a local nash equilibrium},
  author={Heusel, Martin and Ramsauer, Hubert and Unterthiner, Thomas and Nessler, Bernhard and Hochreiter, Sepp},
  journal={Advances in neural information processing systems},
  volume={30},
  year={2017}
}

@inproceedings{InceptionV3,
  title={Rethinking the inception architecture for computer vision},
  author={Szegedy, Christian and Vanhoucke, Vincent and Ioffe, Sergey and Shlens, Jon and Wojna, Zbigniew},
  booktitle={Proceedings of the IEEE conference on computer vision and pattern recognition},
  pages={2818--2826},
  year={2016}
}

@inproceedings{CLIPScore,
  title={CLIPScore: A Reference-free Evaluation Metric for Image Captioning},
  author={Hessel, Jack and Holtzman, Ari and Forbes, Maxwell and Le Bras, Ronan and Choi, Yejin},
  booktitle={Proceedings of the 2021 Conference on Empirical Methods in Natural Language Processing},
  pages={7514--7528},
  year={2021}
}

@article{tsne,
  title={Visualizing data using t-SNE},
  author={Maaten, Laurens van der and Hinton, Geoffrey},
  journal={Journal of machine learning research},
  volume={9},
  number={Nov},
  pages={2579--2605},
  year={2008}
}

@article{Magicmirror,
  title={Magic mirror: Id-preserved video generation in video diffusion transformers},
  author={Zhang, Yuechen and Liu, Yaoyang and Xia, Bin and Peng, Bohao and Yan, Zexin and Lo, Eric and Jia, Jiaya},
  journal={arXiv preprint arXiv:2501.03931},
  year={2025}
}

@article{cpc,
  title={Representation learning with contrastive predictive coding},
  author={Oord, Aaron van den and Li, Yazhe and Vinyals, Oriol},
  journal={arXiv preprint arXiv:1807.03748},
  year={2018}
}

@article{Concat-ID,
  title={Concat-ID: Towards Universal Identity-Preserving Video Synthesis},
  author={Zhong, Yong and Yang, Zhuoyi and Teng, Jiayan and Gu, Xiaotao and Li, Chongxuan},
  journal={arXiv preprint arXiv:2503.14151},
  year={2025}
}

@article{Skyreels-a2,
  title={Skyreels-a2: Compose anything in video diffusion transformers},
  author={Fei, Zhengcong and Li, Debang and Qiu, Di and Wang, Jiahua and Dou, Yikun and Wang, Rui and Xu, Jingtao and Fan, Mingyuan and Chen, Guibin and Li, Yang and others},
  journal={arXiv preprint arXiv:2504.02436},
  year={2025}
}

@inproceedings{MAGREF,
title={{MAGREF}: Masked Guidance for Any-Reference Video Generation with Subject Disentanglement},
author={Yufan Deng and Yuanyang Yin and Xun Guo and Yizhi Wang and Jacob Zhiyuan Fang and Shenghai Yuan and Yiding Yang and Angtian Wang and Bo Liu and Haibin Huang and Chongyang Ma},
booktitle={The Fourteenth International Conference on Learning Representations},
year={2026}
}

@inproceedings{Stand-In,
  title={Stand-in: A lightweight and plug-and-play identity control for video generation},
  author={Xue, Bowen and Duan, Zheng-Peng and Yan, Qixin and Wang, Wenjing and Liu, Hao and Guo, Chun-Le and Li, Chongyi and Li, Chen and Lyu, Jing},
  booktitle={Proceedings of the IEEE/CVF conference on computer vision and pattern recognition},
  year={2026}
}

@inproceedings{jin2025latexblend,
  title={Latexblend: Scaling multi-concept customized generation with latent textual blending},
  author={Jin, Jian and Yu, Zhenbo and Shen, Yang and Fu, Zhenyong and Yang, Jian},
  booktitle={2025 IEEE/CVF Conference on Computer Vision and Pattern Recognition (CVPR)},
  pages={23585--23594},
  year={2025},
  organization={IEEE}
}

@article{jin2026latent,
  title={Latent Textual Space: Pivot for Scaling Customized Text-to-Image and Video Generation},
  author={Jin, Jian and Yu, Zhenbo and Shen, Yang and Fu, Zhenyong and Yang, Jian},
  journal={IEEE Transactions on Pattern Analysis and Machine Intelligence},
  year={2026},
  publisher={IEEE}
}

@InProceedings{IPA,
  title = 	 {Implicit Preference Alignment for Human Image Animation},
  author =       {Wang, Yuanzhi and Ren, Xuhua and Cheng, Jiaxiang and Ma, Bing and Yu, Kai and Zheng, Tianxiang and Lu, Qinglin and Cui, Zhen},
  booktitle = 	 {Proceedings of the 43rd International Conference on Machine Learning},
  pages = 	 {129632--129651},
  year = 	 {2026},
  volume = 	 {306},
  publisher =    {PMLR}
}

@article{IMPFR,
  title={Incomplete Multimodal Probability Flow Recovery for Emotion Recognition},
  author={Wang, Yuanzhi and Cui, Zhen and Liu, Mengyi and Zhang, Xiaoya and Li, Zechao},
  journal={IEEE Transactions on Pattern Analysis and Machine Intelligence},
  year={2026},
  publisher={IEEE}
}

@article{TraFrCo,
  title={Training-free controllable text-guided video editing},
  author={Wang, Yuanzhi and Li, Yong and Liu, Mengyi and Zhang, Xiaoya and Cui, Zhen and Yang, Jian},
  journal={IEEE Transactions on Circuits and Systems for Video Technology},
  year={2026},
  publisher={IEEE}
}

@inproceedings{ReAtCo,
  title={Re-attentional controllable video diffusion editing},
  author={Wang, Yuanzhi and Li, Yong and Liu, Mengyi and Zhang, Xiaoya and Liu, Xin and Cui, Zhen and Chan, Antoni B},
  booktitle={Proceedings of the AAAI Conference on Artificial Intelligence},
  volume={39},
  number={8},
  pages={8123--8131},
  year={2025}
}

@article{TCVE,
  title={Edit temporal-consistent videos with image diffusion model},
  author={Wang, Yuanzhi and Li, Yong and Zhang, Xiaoya and Liu, Xin and Dai, Anbo and Chan, Antoni B and Cui, Zhen},
  journal={ACM Transactions on Multimedia Computing, Communications and Applications},
  volume={20},
  number={12},
  pages={1--16},
  year={2024},
  publisher={ACM New York, NY}
}

@article{IMDer,
  title={Incomplete multimodality-diffused emotion recognition},
  author={Wang, Yuanzhi and Li, Yong and Cui, Zhen},
  journal={Advances in Neural Information Processing Systems},
  volume={36},
  pages={17117--17128},
  year={2023}
}

@inproceedings{DiCMoR,
  title={Distribution-consistent modal recovering for incomplete multimodal learning},
  author={Wang, Yuanzhi and Cui, Zhen and Li, Yong},
  booktitle={2023 IEEE/CVF International Conference on Computer Vision (ICCV)},
  pages={21968--21977},
  year={2023},
  organization={IEEE}
}

@inproceedings{Openhumanvid,
  title={Openhumanvid: A large-scale high-quality dataset for enhancing human-centric video generation},
  author={Li, Hui and Xu, Mingwang and Zhan, Yun and Mu, Shan and Li, Jiaye and Cheng, Kaihui and Chen, Yuxuan and Chen, Tan and Ye, Mao and Wang, Jingdong and others},
  booktitle={2025 IEEE/CVF Conference on Computer Vision and Pattern Recognition (CVPR)},
  pages={7752--7762},
  year={2025},
  organization={IEEE}
}
}


\appendix

\section{Appendix}

\subsection{Observations in Learned Dictionary}
\begin{figure*}[h]
\centering{\includegraphics[width=\linewidth]{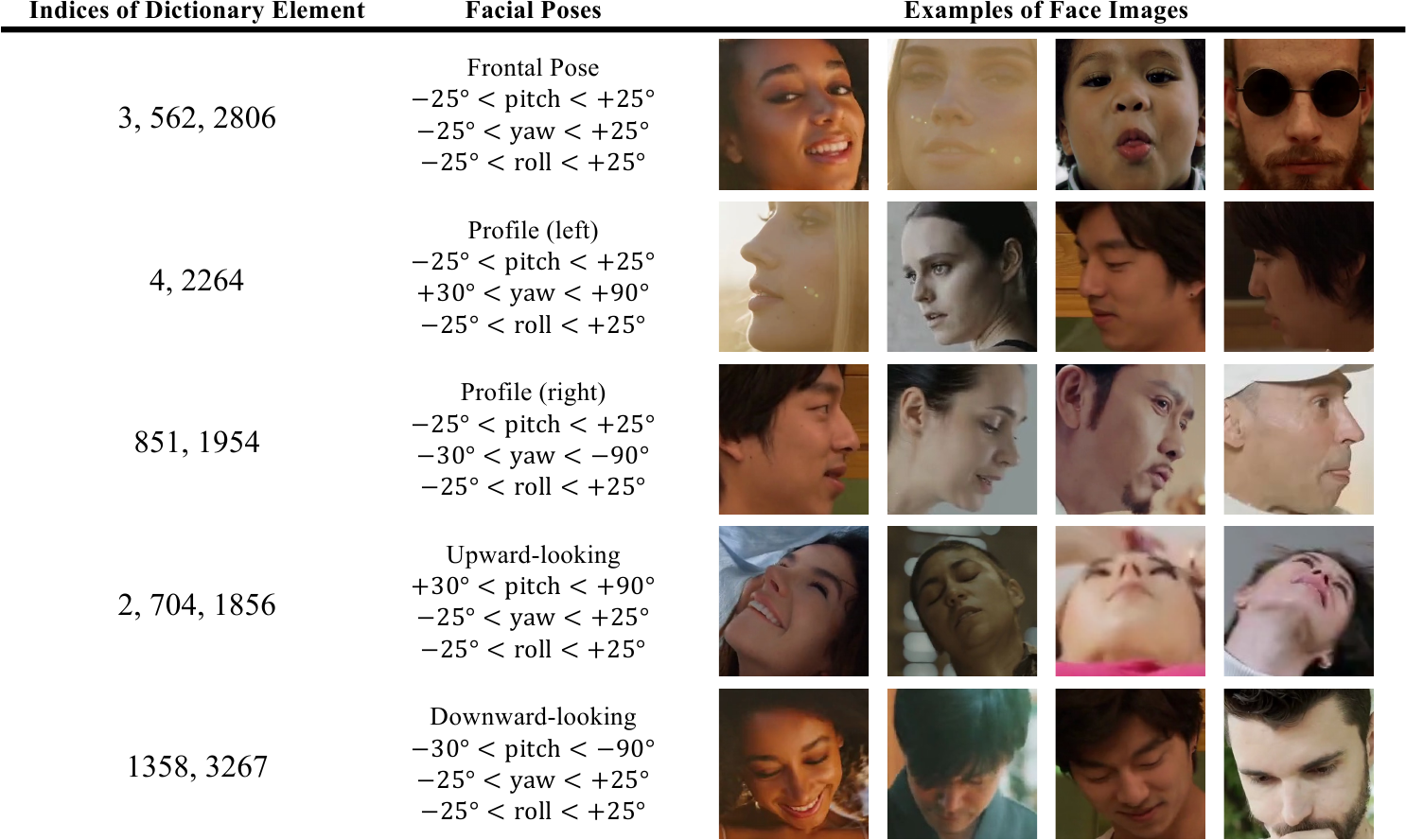}}
	\caption{Visualization of learned pose-shared dictionary. We visualize five representative facial poses and observe that similar poses tend to frequently activate particular dictionary elements.
	}
	\label{fig:dir_stat}
\end{figure*}

To explicitly observe the learned pose-shared dictionary, we visualize the activations of the dictionary for five representative facial poses in Fig.~\ref{fig:dir_stat}.
Specifically, we screen all face images of five facial poses in our dataset based on Euler angles and feed them into the pose-shared identity aligner to calculate the dictionary weight corresponding to each face image.
We then record the indices of the top-10 elements in each weight vector, representing the most prominently activated dictionary elements for each face image.
From the results presented in Fig.~\ref{fig:dir_stat}, we observe that particular dictionary elements are consistently activated by face images with similar poses.
For example, the \textit{frontal pose} tends to activate dictionary elements with indices 3, 562, and 2806, whereas the \textit{upward-looking pose} frequently activates those with indices 2, 704, and 1856.
This observation demonstrates that the learned dictionary captures meaningful patterns, potentially enabling robust representations of faces exhibiting a wide range of poses.

\subsection{Exploring the Effects of Different Dictionary Elements}\label{dic_num}

\begin{figure*}[h]
\centering{\includegraphics[width=\linewidth]{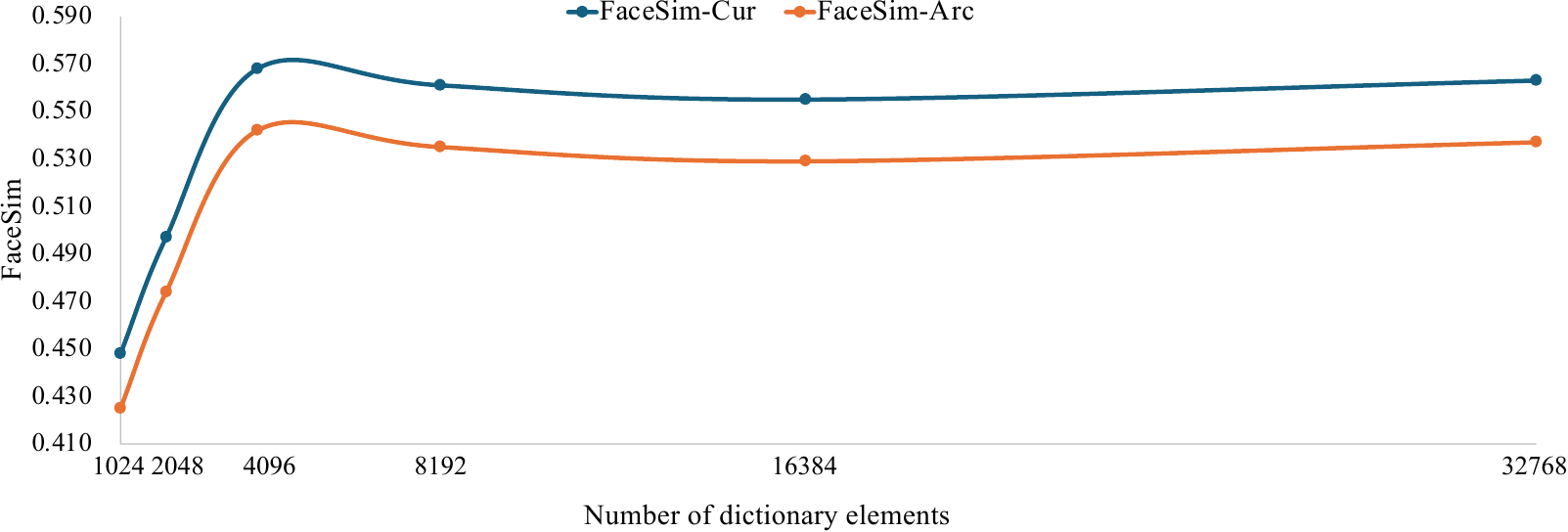}}
	\caption{Ablation study on different dictionary elements.
	}
	\label{fig:num_dir}
\end{figure*}

In this section, we conduct the ablation study to explore the effects of different dictionary elements in $\mathbf{D}$.
Specifically, we conduct six sets of experiments in which the number of dictionary elements is set to 1024, 2048, 4096, 8192, 16384, and 32768, respectively.
Fig.~\ref{fig:num_dir} illustrates the performance of our FaithfulFaces with various dictionary elements under two FaceSim metrics, and we can observe that the best performance is reached when the number of dictionary elements is set to 4096.
Subsequently, the performance exhibits only slight variations as the number of dictionary elements increases further.
In addition, the quantitative results of different dictionary elements are listed in Tab.~\ref{tab:num_dic}.
Hence, we use 4096 dictionary elements in our experiments.

\begin{table}[h]
	\centering
	\caption{Quantitative results of different dictionary elements.}\label{tab:num_dic}
	\scalebox{1.0}{
         \begin{tabular}{c|c|c|c|c}
        \hline
         Dictionary Elements & FaceSim-Cur~$\uparrow$ &  FaceSim-Arc~$\uparrow$ & FID~$\downarrow$ & CLIPScore~$\uparrow$ \\
        \hline
        \hline
     1024 & 0.448 & 0.425 & 183.39 & 32.28 \\
     2048 & 0.497& 0.474 & 176.80 & 32.89 \\
     4096 & \textbf{0.568} & \textbf{0.542} & \textbf{164.24} & \textbf{33.93} \\
     8192 & 0.561 & 0.535 & 165.94 & 33.82 \\
     16384 & 0.555 & 0.529 & 166.64 & 33.83 \\
     32768 & 0.563 & 0.537 & 165.96 & 33.82 \\
        \hline
	\end{tabular}}
\end{table}

\subsection{Qualitative Analysis of Ablation Study for Key Components}\label{vis-abl}
\begin{figure*}[h]
\centering{\includegraphics[width=\linewidth]{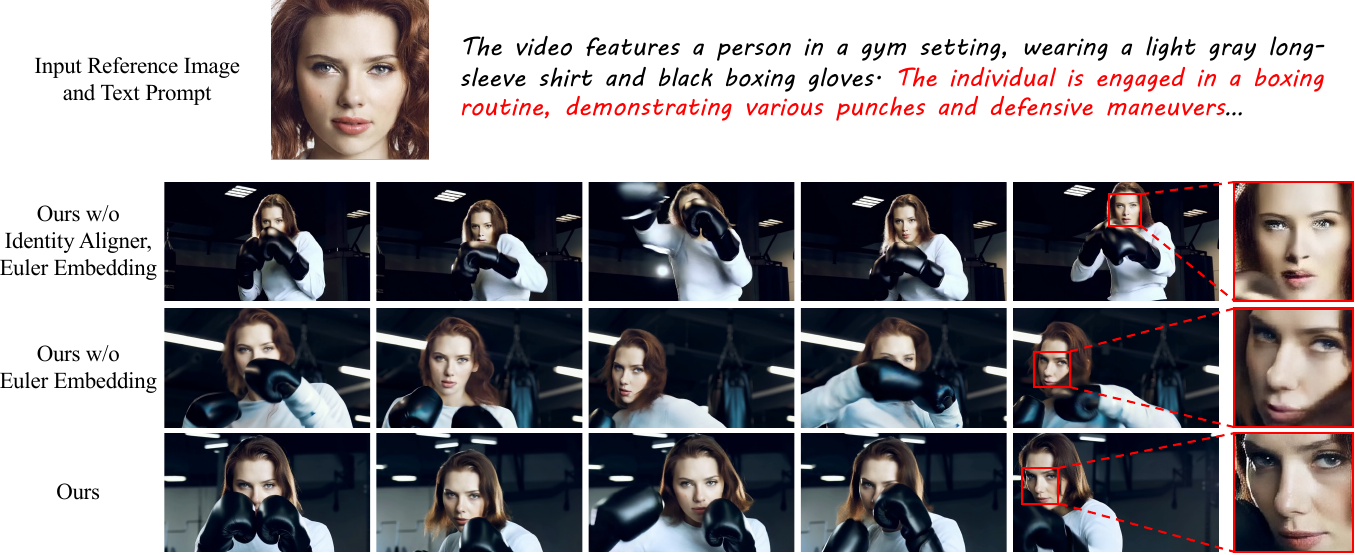}}
	\caption{Visual comparisons of the ablation study for key components in FaithfulFaces.
	}
	\label{fig:vis-ablation}
\end{figure*}

In addition to the quantitative results of the ablation study presented in Tab.~\ref{tab:Ablation}, we provide a qualitative analysis with visualizations in Fig.~\ref{fig:vis-ablation}. We can observe that \textit{Ours w/o (Identity Aligner, Euler Embedding)} shows obvious distortion of facial structures and facial details due to the lack of global facial pose awareness.
\textit{Ours w/o Euler Embedding} mitigates facial distortions owing to the global facial pose representation provided by our pose-shared identity aligner; however, its identity consistency and facial stability remain suboptimal.
In contrast, \textit{Ours} generates high-quality results with clear facial structures and consistent identity details.

\subsection{Ablation Study of Pooling Operation Type}\label{abl_pooling}
We evaluate the effects of different pooling operation types in the pose-shared identity aligner, and the results are listed in Tab.~\ref{tab:pooling}.
From these results, we can observe that performance is optimal when using the max pooling operation.
We also analyzed the potential underlying reason: the majority of information in face images of the same identity across different poses is similar or redundant. 
Therefore, using max pooling can alleviate a large amount of redundant information and extract highly abstract pose variations.

\begin{table}[h]
	\centering
	\caption{Quantitative results of different pooling operation types.}\label{tab:pooling}
	\scalebox{1.0}{
         \begin{tabular}{c|c|c|c|c}
        \hline
         Pooling Type & FaceSim-Cur~$\uparrow$ &  FaceSim-Arc~$\uparrow$ & FID~$\downarrow$ & CLIPScore~$\uparrow$ \\
        \hline
        \hline
     Sum Pooling & 0.444 & 0.421 & 185.40 & 32.17 \\
     Mean Pooling & 0.559 & 0.533 & 165.64 & 33.84 \\
     Max Pooling & \textbf{0.568} & \textbf{0.542} & \textbf{164.24} & \textbf{33.93} \\
        \hline
	\end{tabular}}
\end{table}

\subsection{Ablation Study of Different Identity Features}\label{identity-features}
We conduct the ablation study by replacing the aligner features with the ArcFace~\cite{Arc} feature and the CLIP~\cite{clip} feature to measure the aligner's specific effect.
The experimental results are listed in Tab.~\ref{tab:features}.
We can observe that performance significantly deteriorates when using the ArcFace feature or the CLIP feature.
The underlying reason is \textbf{i)} ArcFace feature is unable to represent global facial pose information; and \textbf{ii)} CLIP features inherently encode the alignment between text and image modalities, rather than being dedicated to representing facial identity.
This ablation study further demonstrates the effectiveness of the pose-shared identity aligner.

\begin{table}[h]
	\centering
	\caption{Quantitative results of different identity features.}\label{tab:features}
	\scalebox{1.0}{
         \begin{tabular}{c|c|c|c|c}
        \hline
         Feature Type & FaceSim-Cur~$\uparrow$ &  FaceSim-Arc~$\uparrow$ & FID~$\downarrow$ & CLIPScore~$\uparrow$ \\
        \hline
        \hline
     CLIP & 0.447 & 0.422 & 183.57 & 32.13 \\
     ArcFace & 0.475 & 0.453 & 177.20 & 32.70 \\
     Aligner (Ours) & \textbf{0.568} & \textbf{0.542} & \textbf{164.24} & \textbf{33.93} \\
        \hline
	\end{tabular}}
\end{table}

\subsection{Stability of Contrastive Loss in Optimization Procedure}
\begin{figure*}[h]
\centering{\includegraphics[width=\linewidth]{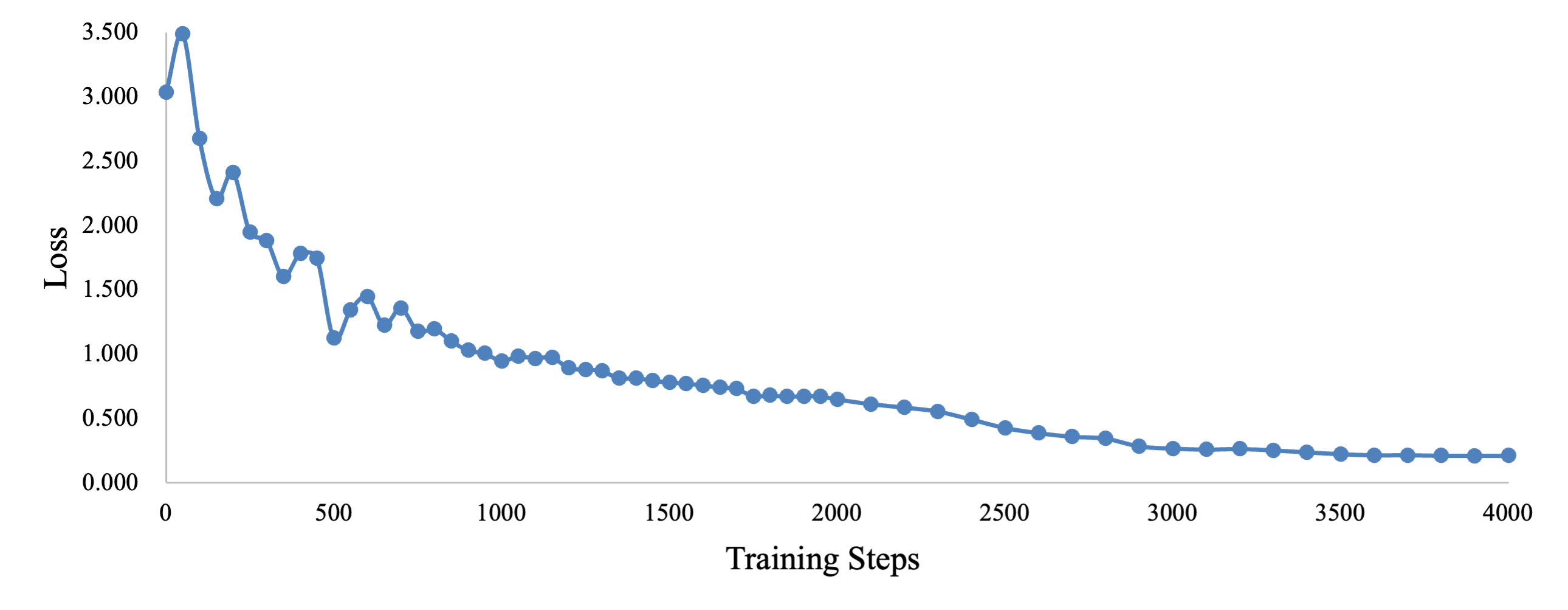}}
	\caption{The value of the contrastive loss in the pose-shared identity aligner (named aligner loss) during the training process.
	}
	\label{fig:loss}
\end{figure*}

Fig.~\ref{fig:loss} illustrates the value of the contrastive loss in the pose-shared identity aligner from different training steps.
We can observe that the loss value gradually decreases during training and eventually converges to approximately 0.2.
These results demonstrate the stability and convergence of contrastive learning for the pose-shared identity aligner during training.

\subsection{Discussion on Non-frontal View Robustness}\label{non-frontal-discuss}
In this section, we discuss the robustness of the method when the input reference image is a non-frontal view.
Specifically, we collect 10 identities with both frontal and non-frontal face images for ablation and comparative experiments (the strongest baseline Phantom~\cite{Phantom} as the representative). 
The results are shown in the Tab.~\ref{tab:nonfrontal}, we can observe that both Ours w/o Aligner and the strongest baseline Phantom suffer a severe performance decrease exceeding 50\% when non-frontal face images are used as input. 
In contrast, our method can control the performance decrease within 25\%. 
These results provide strong evidence that the pose-shared identity aligner can mitigate performance degradation in non-frontal face scenarios.

\begin{figure*}[h]
\centering{\includegraphics[width=\linewidth]{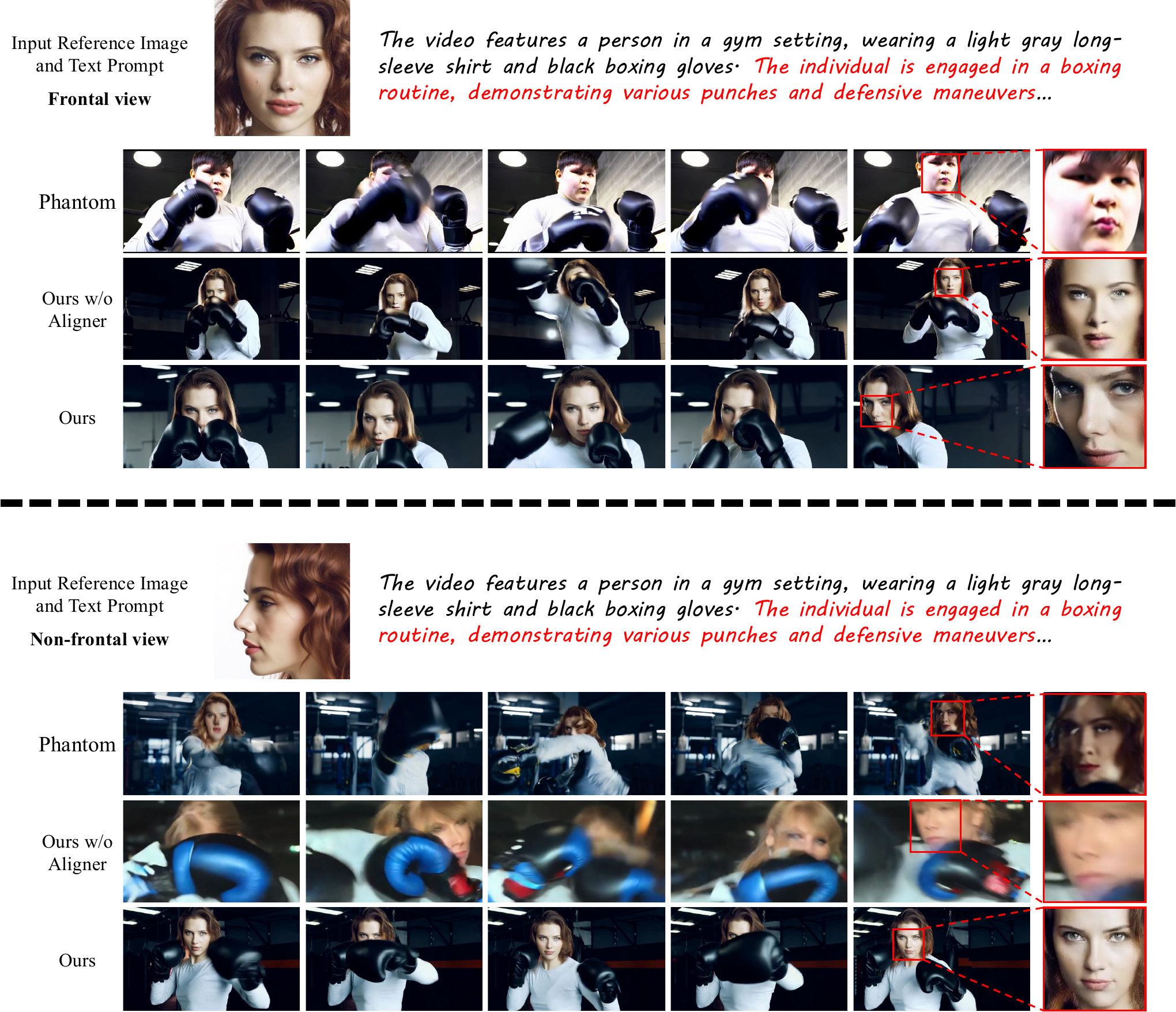}}
	\caption{Visual comparisons of frontal view and non-frontal view. We can observe that when using a non-frontal image as input, the faces generated by the SOTA method Phantom and the baseline method (Ours w/o Aligner) completely collapse. In contrast, our method still maintains identity consistency.
	}
	\label{fig:nonfrontal}
\end{figure*}

Additionally, we further provide visualization results of the frontal view and the non-frontal view in Fig.~\ref{fig:nonfrontal}.
We can further observe that when using a non-frontal image as input, the faces generated by the Phantom and Ours w/o Aligner completely collapse. 
In contrast, our method still maintains identity consistency.
These visual results once again demonstrate that our method is capable of improving identity consistency in non-frontal face scenarios.

\begin{table}[h]
	\centering
	\caption{Quantitative results of frontal view and non-frontal view. The values reported in each cell denote FaceSim-Cur/ FaceSim-Arc}\label{tab:nonfrontal}
	\scalebox{1.0}{
         \begin{tabular}{c|c|c}
        \hline
         Methods & Frontal &  Non-frontal \\
        \hline
        \hline
     Phantom~\cite{Phantom} & 0.470 / 0.435 & 0.231 ($\downarrow$ 50.85\%) / 0.216 ($\downarrow$ 50.34\%)  \\
     Ours w/o Aligner & 0.448 / 0.423 & 0.202 ($\downarrow$ 54.91\%) / 0.197 ($\downarrow$ 53.43\%)   \\
     Ours & 0.544 / 0.522 & 0.409 ($\downarrow$ 24.82\%) / 0.392 ($\downarrow$ 24.90\%)  \\
        \hline
	\end{tabular}}
\end{table}

\subsection{Discussion on the Robustness of Identity Aligner for Euler Angles}\label{aligner-euler-discuss}
In our pose-shared identity aligner, the sparsity design of the dictionary representation mechanism inherently possesses a certain degree of noise tolerance.
To demonstrate this, we conduct the ablation experiments involving Euler angle perturbations.
Specifically, given four perturbation ranges for Euler angles: $-5^\circ \sim +5^\circ$, $-10^\circ \sim +10^\circ$, $-15^\circ \sim +15^\circ$, $-20^\circ \sim +20^\circ$.
We apply random perturbations within these ranges to the predicted Euler angles to observe the impact on performance. 
The experimental results are shown in the Tab.~\ref{tab:euler_errors}, we can observe that performance exhibits only minor variations within the perturbation range of $-15^\circ \sim +15^\circ$. 
Significant performance degradation occurs only when perturbations exceed $15^\circ$ (representing substantial errors). 
These results demonstrate that our method exhibits high robustness to Euler angle errors within a certain range.

\begin{table}[h]
	\centering
	\caption{Quantitative results under different Euler angle perturbations.}\label{tab:euler_errors}
	\scalebox{1.0}{
         \begin{tabular}{c|c|c|c|c}
        \hline
         Perturbation Range & FaceSim-Cur~$\uparrow$ &  FaceSim-Arc~$\uparrow$ & FID~$\downarrow$ & CLIPScore~$\uparrow$ \\
        \hline
        \hline
     No Perturbation & 0.568 & 0.542 & 164.24 & 33.93 \\
     $-5^\circ \sim +5^\circ$ & 0.566& 0.540 & 164.54 & 33.89  \\
     $-10^\circ \sim +10^\circ$ & 0.557 &0.531 & 166.06 & 33.77   \\
     $-15^\circ \sim +15^\circ$ & 0.552 & 0.526 & 167.07 & 33.74 \\
     $-20^\circ \sim +20^\circ$ & 0.523 & 0.499 & 172.54 & 33.56 \\
        \hline
	\end{tabular}}
\end{table}

\subsection{Pose-Conditioned Identity Preservation Analysis}
To further evaluate the effectiveness of FaithfulFaces in preserving facial identity under large pose variations, we conduct a pose-conditioned identity analysis. Specifically, we estimate the yaw angle of each generated frame with a valid face detection and group the frames into six $30^\circ$ intervals spanning $-90^\circ$ to $90^\circ$. For each interval, we compute the average FaceSim-Cur and FaceSim-Arc scores between the generated faces and their corresponding reference identities. This evaluation directly measures identity preservation under different generated head poses without requiring ground-truth pose trajectories.
As shown in Table~\ref{tab:pose_analysis}, identity similarity generally decreases for all methods as the head pose approaches extreme profile views. Nevertheless, FaithfulFaces consistently achieves the highest FaceSim-Cur and FaceSim-Arc scores across all pose intervals. Notably, in the challenging $[-90^\circ,-60^\circ)$ interval, FaithfulFaces achieves scores of 0.466 and 0.448, substantially outperforming the strongest competing baseline, Phantom, which obtains 0.253 and 0.244, respectively. Similar improvements are observed for large positive yaw angles. These results demonstrate that the proposed pose-shared identity alignment effectively mitigates identity degradation under substantial head rotations, supporting more robust identity preservation across diverse viewing angles.

\begin{table}[h]
    \centering
    \caption{Pose-conditioned identity preservation across different yaw intervals. Each entry reports FaceSim-Cur / FaceSim-Arc ($\uparrow$). The best results are highlighted in bold.}
    \label{tab:pose_analysis}
    \begin{tabular}{c|c|c|c|c}
        \hline
        Yaw Interval & VACE-14B & Phantom & MAGREF & \textbf{FaithfulFaces} \\
        \hline
        \hline
        $[-90^\circ,-60^\circ)$
        & 0.232/0.213 & 0.253/0.244 & 0.245/0.226 & \textbf{0.466/0.448} \\
        $[-60^\circ,-30^\circ)$
        & 0.355/0.332 & 0.382/0.363 & 0.363/0.340 & \textbf{0.571/0.546} \\
        $[-30^\circ,0^\circ)$
        & 0.471/0.446 & 0.502/0.476 & 0.484/0.458 & \textbf{0.644/0.623} \\
        $[0^\circ,30^\circ)$
        & 0.476/0.452 & 0.509/0.485 & 0.489/0.461 & \textbf{0.651/0.624} \\
        $[30^\circ,60^\circ)$
        & 0.363/0.341 & 0.392/0.374 & 0.371/0.350 & \textbf{0.575/0.554} \\
        $[60^\circ,90^\circ]$
        & 0.231/0.216 & 0.263/0.251 & 0.242/0.229 & \textbf{0.473/0.452} \\
        \hline
    \end{tabular}
\end{table}

\subsection{Effectiveness of the Pose-shared Dictionary}
To investigate the necessity of the proposed pose-shared dictionary, we consider a straightforward alternative that directly concatenates facial tokens with Euler angle embeddings and injects them into the video generation backbone, without dictionary-based aggregation. Both variants adopt the same backbone, training data, and optimization settings for a controlled comparison. As shown in Tab.~\ref{tab:dictionary_ablation}, removing the dictionary substantially degrades performance across all metrics. Specifically, the full model improves FaceSim-Cur and FaceSim-Arc from 0.445 and 0.426 to 0.568 and 0.542, respectively, while reducing FID from 182.95 to 164.24. These improvements suggest that directly injecting pose-conditioned facial features is insufficient for effective identity preservation, as the features may still contain pose-specific and redundant information. In contrast, the proposed dictionary provides learnable shared bases to aggregate identity-relevant features across different poses, thereby facilitating more discriminative and pose-robust identity representations.

\begin{table}[h]
    \centering
    \caption{Ablation study on the effectiveness of the pose-shared dictionary.}
    \label{tab:dictionary_ablation}
    \begin{tabular}{c|c|c|c|c}
        \hline
        Method & FaceSim-Cur $\uparrow$ & FaceSim-Arc $\uparrow$ & FID $\downarrow$ & CLIPScore $\uparrow$ \\
        \hline
        \hline
        Ours w/o Dictionary & 0.445 & 0.426 & 182.95 & 32.14 \\
        \textbf{Ours} & \textbf{0.568} & \textbf{0.542} & \textbf{164.24} & \textbf{33.93} \\
        \hline
    \end{tabular}
\end{table}

\subsection{Video-Level Quality Evaluation}
To further assess the temporal consistency and motion quality of generated videos, we conduct additional evaluations using three VBench metrics: Subject Consistency, Motion Smoothness, and Dynamic Degree. As shown in Tab.~\ref{tab:vbench}, FaithfulFaces consistently achieves the best performance across all three metrics. These results indicate that FaithfulFaces not only enhances facial identity preservation, as demonstrated by the identity similarity metrics, but also maintains strong video-level consistency and smooth motion dynamics.

\begin{table}[h]
    \centering
    \caption{Video-level quality evaluation using VBench metrics. Higher values indicate better performance.}
    \label{tab:vbench}
    \begin{tabular}{c|c|c|c}
        \hline
        Method & Subject Consistency $\uparrow$ & Motion Smoothness $\uparrow$ & Dynamic Degree $\uparrow$ \\
        \hline
        \hline
        VACE-14B & 95.80\% & 96.69\% & 97.60\% \\
        Phantom & 96.34\% & 97.07\% & 97.22\% \\
        MAGREF & 95.92\% & 96.92\% & 98.33\% \\
        \textbf{FaithfulFaces} & \textbf{98.33\%} & \textbf{97.99\%} & \textbf{98.61\%} \\
        \hline
    \end{tabular}
\end{table}

\subsection{Human Preference Study}
To further assess the perceptual quality of generated videos beyond automated metrics, we conduct a human preference study involving 10 participants and 30 sets of generated videos. Each set includes videos generated by FaithfulFaces and three representative baselines, i.e., VACE, Phantom, and MAGREF. For each pairwise comparison, participants are presented with the text prompt, reference image, and two generated videos (ours and a baseline) displayed side by side in randomized order.
Participants are asked to select the preferred video by considering three aspects: (1) \textit{Identity Preservation}, assessing the consistency of the generated facial identity with the reference image; (2) \textit{Visual Quality}, evaluating the overall perceptual quality of the generated video; and (3) \textit{Text Alignment}, measuring how well the generated content follows the text prompt. The overall pairwise preference results are reported in Tab.~\ref{tab:human_study}.

\begin{table}[h]
    \centering
    \caption{Human preference study.}
    \label{tab:human_study}
    \begin{tabular}{l|c|c}
        \hline
        Comparison & Ours Win (\%) & Ours Lose (\%) \\
        \hline
        \hline
        Ours vs. VACE    & \textbf{81.3} & 18.7 \\
        Ours vs. Phantom & \textbf{84.7} & 15.3 \\
        Ours vs. MAGREF  & \textbf{78.3} & 21.7 \\
        \hline
    \end{tabular}
\end{table}

\subsection{Implementation Details of the Pose-shared Identity Aligner}
\noindent\textbf{Face Feature Encoding.}
We first detect the facial region using InsightFace and resize the cropped face while preserving its aspect ratio, followed by padding to a resolution of $480 \times 832$. Given a reference face tensor of shape $B \times 3 \times 1 \times 480 \times 832$, the pretrained Wan VAE encodes it into a latent representation of shape $B \times 16 \times 1 \times 60 \times 104$. We then reuse the DiT patch-embedding layer, implemented as a 3D convolution with kernel size and stride $(1,2,2)$, to obtain a feature tensor of shape $B \times 5120 \times 1 \times 30 \times 52$. Flattening the spatiotemporal dimensions produces 1,560 facial tokens, each with 5,120 dimensions, resulting in a tensor of shape $B \times 1560 \times 5120$.

\noindent\textbf{Euler Pose Embeddings.}
We estimate the pitch, yaw, and roll angles using 6DRepNet. Each Euler angle is independently encoded into a 5,120-dimensional sinusoidal embedding and processed by a shared two-layer MLP. The resulting three embeddings are concatenated into a 15,360-dimensional vector and subsequently projected back to 5,120 dimensions. This pose embedding is broadcast along the token dimension and added to the 1,560 facial tokens before dictionary-based aggregation.

\noindent\textbf{Pose-shared Dictionary Aggregation.}
The pose-conditioned facial tokens interact with a learnable dictionary containing 4,096 vectors, each of dimension 5,120. This produces a similarity score tensor of shape $B \times 1560 \times 4096$. We apply max pooling over the 1,560 facial tokens to obtain dictionary-level aggregation weights. A weighted combination of the dictionary vectors then yields a single identity representation of shape $B \times 5120$. Since its feature dimension matches the hidden dimension of VACE-14B, no additional output projection is required.

\noindent\textbf{Identity Conditioning Interface.}
Importantly, the resulting identity representation is not directly added to the raw noisy video latent. Instead, the noisy latent is first processed by the DiT patch-embedding layer and flattened into a video token sequence $\mathbf{X}_t \in \mathbb{R}^{B \times L_t \times 5120}$, where $L_t$ denotes the number of spatiotemporal video tokens. The identity representation $\mathbf{e}_{id} \in \mathbb{R}^{B \times 5120}$ is first reshaped into a single identity token of shape $B \times 1 \times 5120$. We then zero-pad this token along the sequence dimension to match the video token length $L_t$, obtaining $\mathbf{E}_{id} \in \mathbb{R}^{B \times L_t \times 5120}$. The identity-conditioned video tokens are computed as
\begin{equation}
    \widetilde{\mathbf{X}}_t =
    \mathbf{X}_t + \mathbf{E}_{id}.
\end{equation}
Consequently, the identity representation is injected into the first video token before the DiT Transformer blocks, while subsequent self-attention operations propagate identity information across the spatiotemporal sequence. This reshape-and-padding operation ensures dimensional compatibility without modifying the original DiT architecture.

\subsection{Prompt Construction}\label{Prompt}
We now elaborate on the construction of challenging test text prompts designed to drive models to generate videos exhibiting significant facial pose variations, expression changes, and facial occlusions across diverse scenarios.
For character movement, we select several representative scenes: 1) \textbf{Boxing} with facial pose variations and facial occlusions; 2) \textbf{Head shaking and dancing} with facial pose variations; 3) The character \textbf{transitions from having their back to the camera to facing the camera}; 4) \textbf{Ballet} with facial pose variations; 5) \textbf{Speech} with facial pose variations and expression changes; 6) Some descriptions used to generate \textbf{dramatic changes in facial expressions and poses}, as shown in the second case in Fig.~\ref{fig:append_vis_nocom_1}.

With these basic movement scenes, we use GPT-4.1\footnote{https://openai.com/index/gpt-4-1/} to generate information-rich text prompts.
Taking the boxing scene as an example, the generated text prompt is: \textit{``The video features a person in a gym setting, wearing a light gray long-sleeve shirt and black boxing gloves. The individual is engaged in a boxing routine, demonstrating various punches and defensive maneuvers. The camera closely follows the person's movements, keeping their face and gloves prominent in the frame, and capturing detailed facial expressions and dynamic action. The background is dimly lit, with overhead lights providing illumination. The gym environment is evident from the visible equipment and the industrial setting, which adds to the intensity of the scene.''}
Subsequently, based on this, we employ GPT-4.1 to generate text prompts for different background scenarios, such as \textit{``open grassy field''} in Fig.~\ref{fig:vis_1}, \textit{``urban street setting''} in Fig.~\ref{fig:append_vis_nocom_3}.
Ultimately, we curate 20 high-quality text prompts that span various actions and background scenarios.

\subsection{Ethics Statement and Broader Impact}\label{Ethics}
Advancements in identity-preserving text-to-video generation technology are poised to support and empower the creative processes of artists and designers.
FaithfulFaces is capable of generating high-quality, realistic human videos.
However, it also raises concerns regarding misinformation, potentially undermining the reliability of video content.
Additionally, this technology could be misused to generate deceptive content for fraudulent purposes.
It is important to recognize that any technology is susceptible to misuse.
Nevertheless, it is feasible to train a classifier that can distinguish between real and FaithfulFaces-generated videos based on their texture features.

Regarding data ethics, our training data are curated from OpenHumanVid, obtained through its official application-and-review process. To reduce potential privacy and redistribution risks, we do not release the original training videos, extracted facial images, or the processed training dataset.

\noindent\textbf{Responsible Release.}
To facilitate reproducibility, we plan to release the implementation code of FaithfulFaces. However, we will not redistribute the original training videos, extracted facial images, or processed training dataset due to potential privacy and licensing concerns. The public release of trained model weights will be subject to further review of data licensing, privacy implications, and potential misuse risks. We do not currently commit to providing a public inference API.

\subsection{Reproducibility Statement}
First, we have explained the implementation of  FaithfulFaces in detail in Sec.~\ref{Implementation}.
Second, we have explained the details of training and inference in Fig.~\ref{fig:2} and Sec.~\ref{Overview}.
Third, we have explained the details of the dataset construction in Sec.~\ref{pipeline}.
Finally, we plan to release the implementation code of FaithfulFaces to facilitate reproducibility, while the training data will not be redistributed due to privacy and licensing considerations.

\subsection{The Use of Large Language Models}
This submission utilizes a large language model for grammar checking.

\subsection{More Visualization Results}\label{app_vis}
\begin{figure*}[!ht]
\centering{\includegraphics[width=\linewidth]{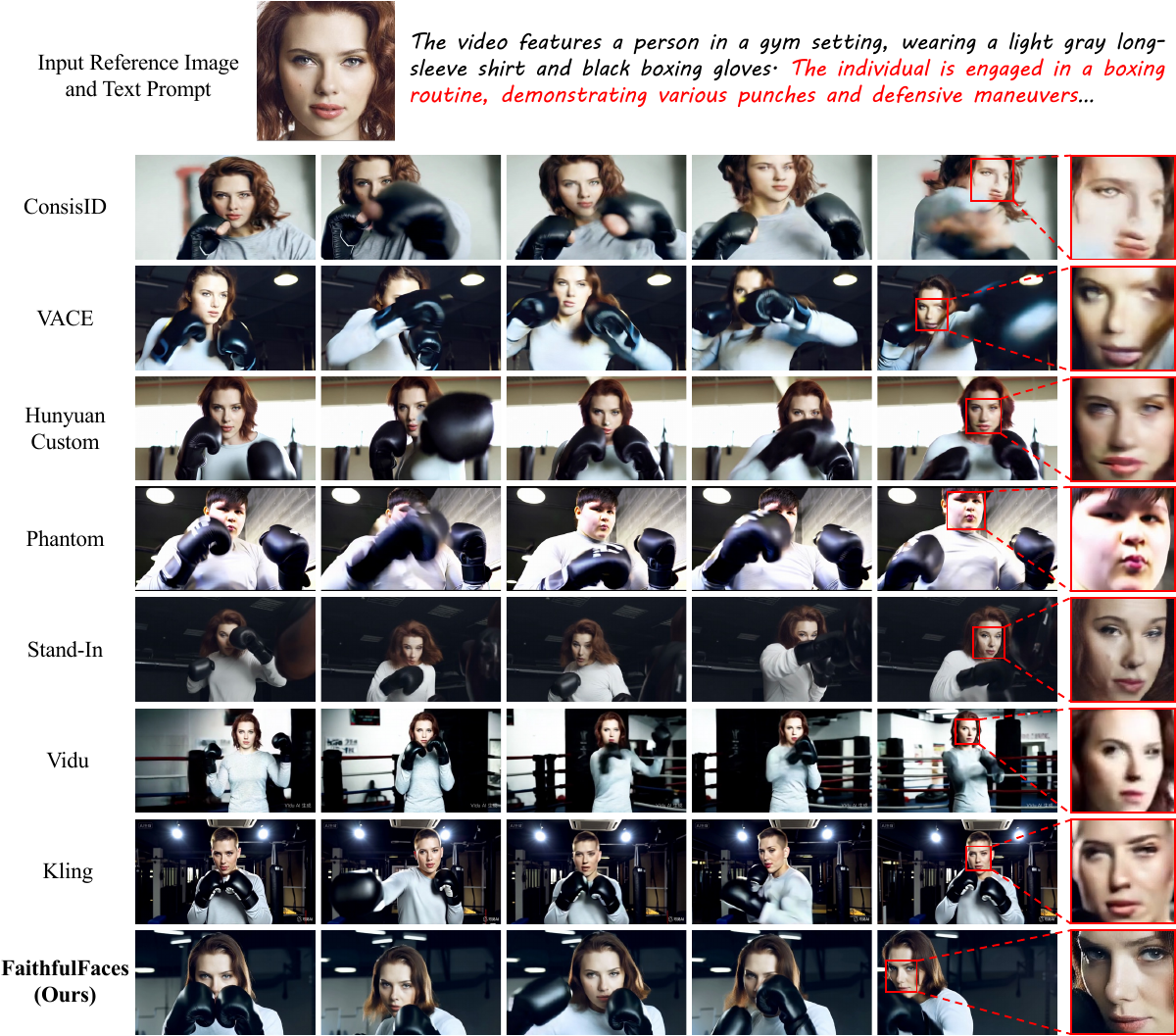}}
	\caption{Complete visual comparisons of different methods for the case of Fig.~\ref{fig:1}.
	}
	\label{fig:append_vis_1}
\end{figure*}

\begin{figure*}[!ht]
\centering{\includegraphics[width=\linewidth]{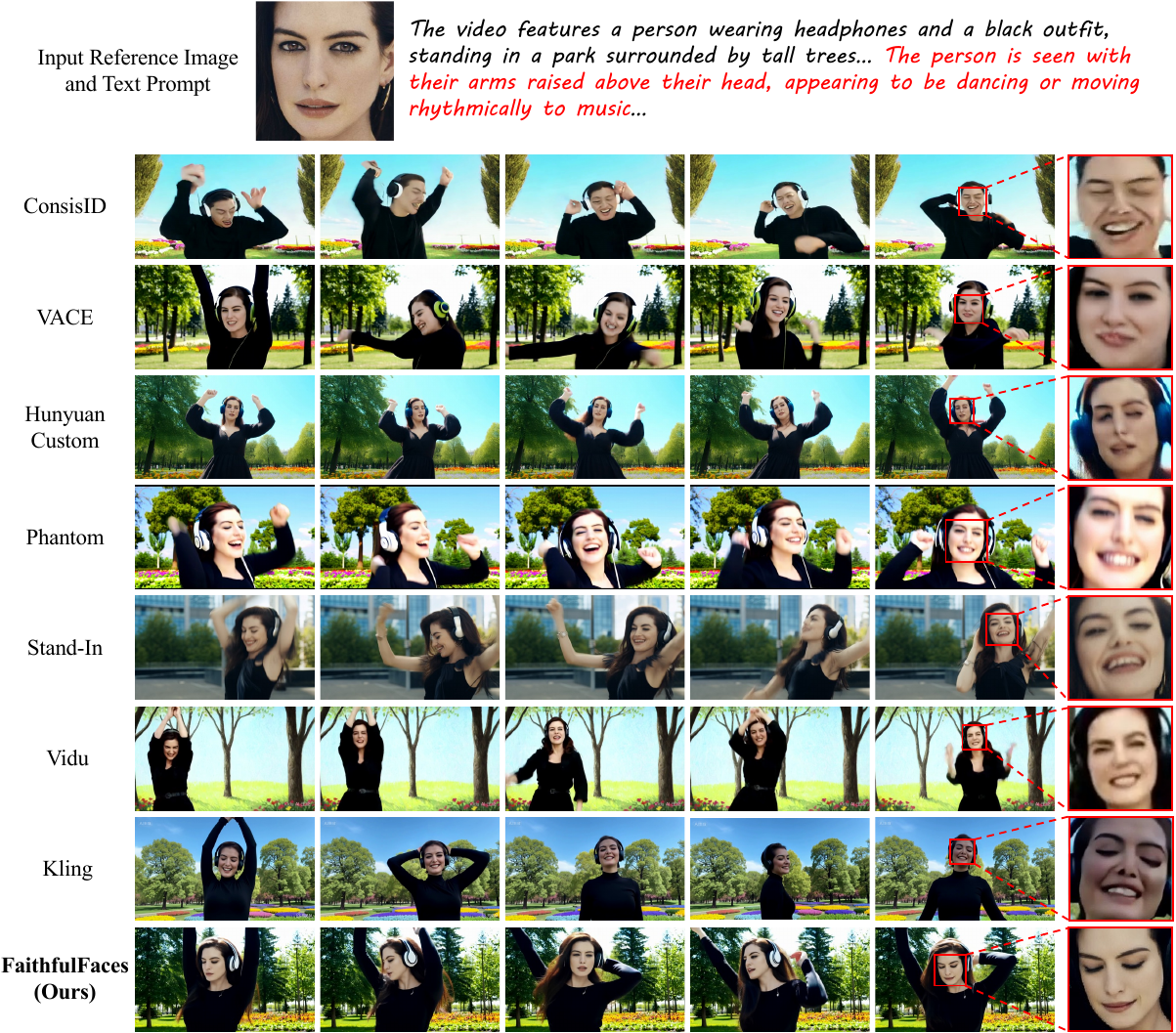}}
	\caption{More visual comparisons of different methods.
	}
	\label{fig:append_vis_2}
\end{figure*}

\begin{figure*}[!ht]
\centering{\includegraphics[width=\linewidth]{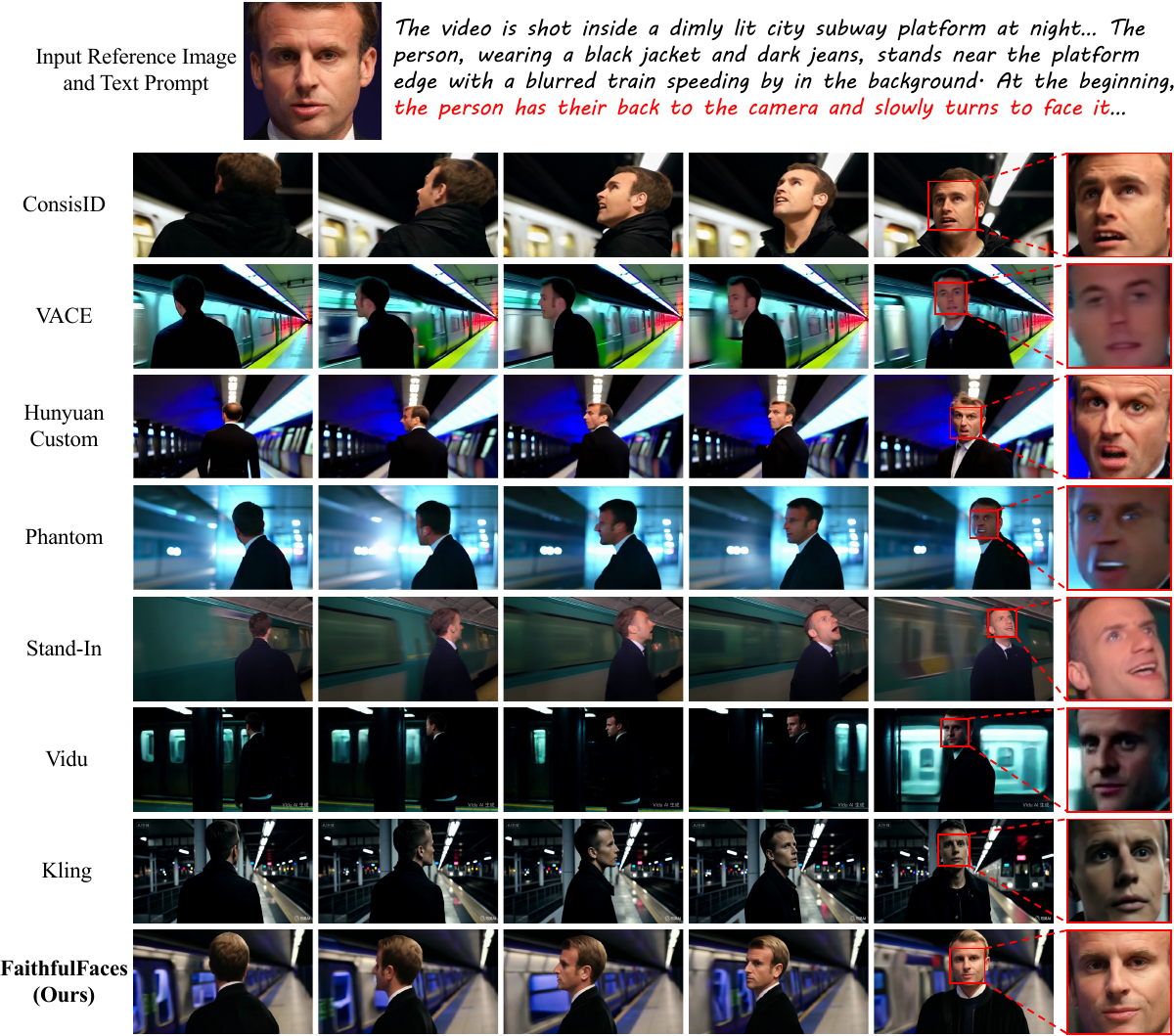}}
	\caption{More visual comparisons of different methods.
	}
	\label{fig:vis_2}
\end{figure*}

\begin{figure*}[!ht]
\centering{\includegraphics[width=\linewidth]{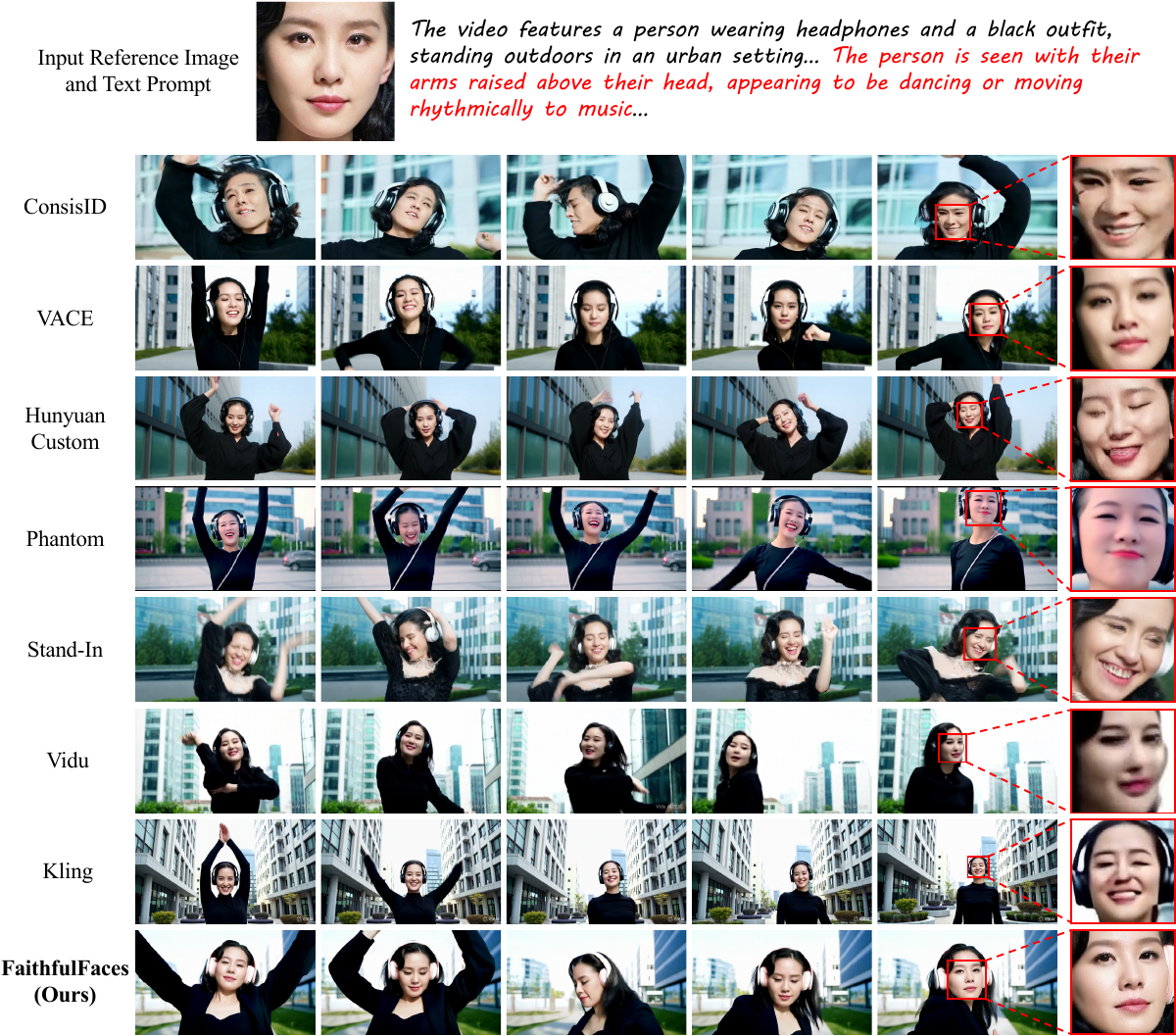}}
	\caption{More visual comparisons of different methods.
	}
	\label{fig:append_vis_3}
\end{figure*}

\begin{figure*}[!ht]
\centering{\includegraphics[width=\linewidth]{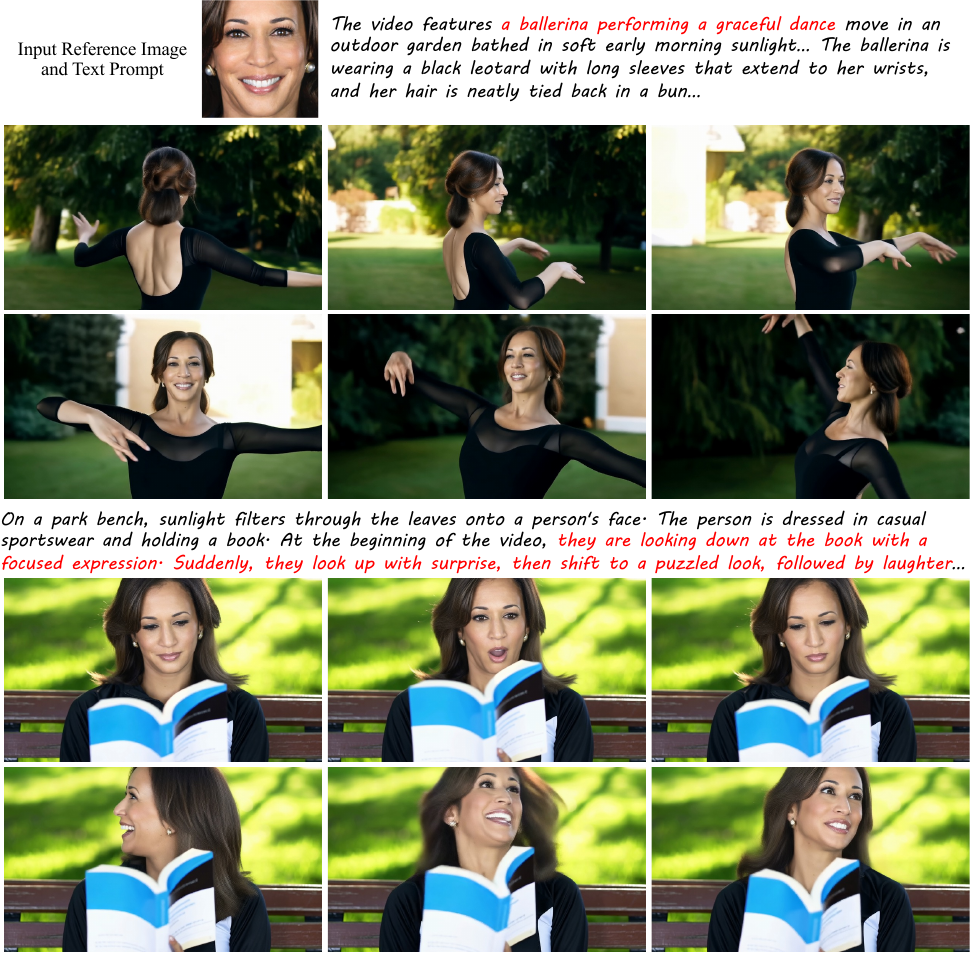}}
	\caption{More showcases of identity-preserving videos generated by our FaithfulFaces.
	}
	\label{fig:append_vis_nocom_1}
\end{figure*}

\begin{figure*}[!ht]
\centering{\includegraphics[width=\linewidth]{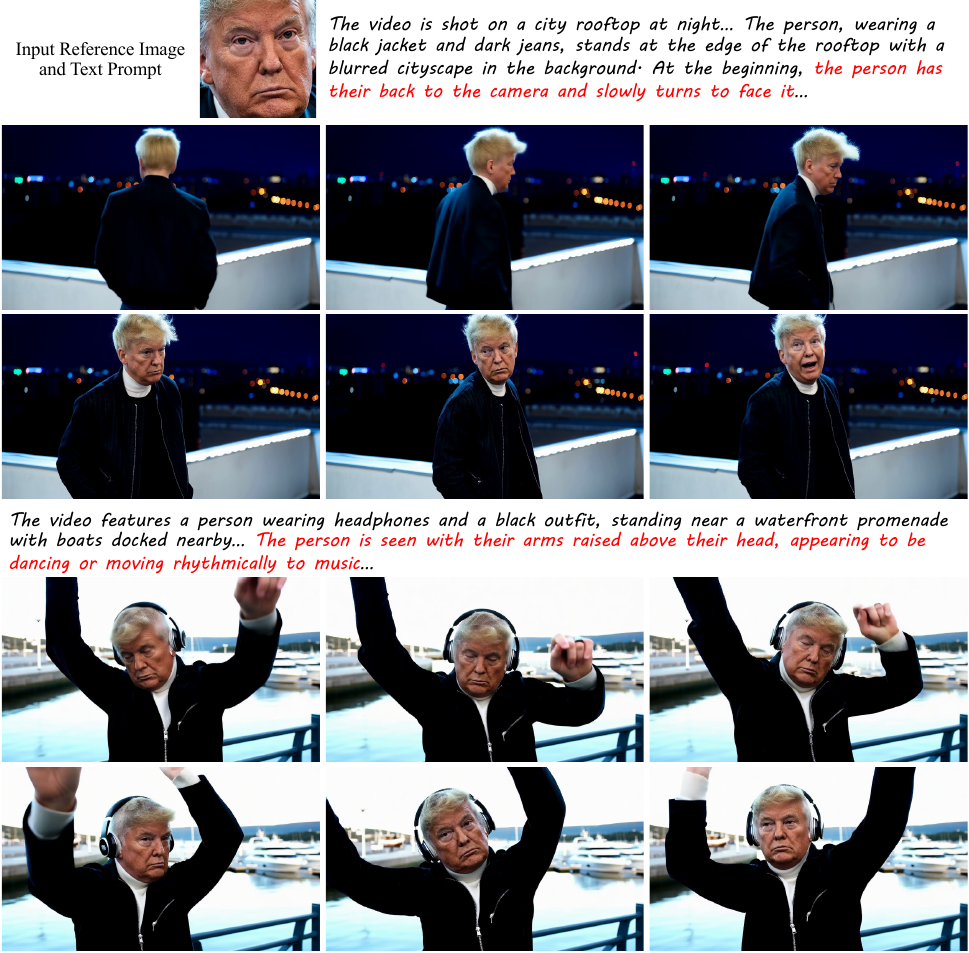}}
	\caption{More showcases of identity-preserving videos generated by our FaithfulFaces.
	}
	\label{fig:append_vis_nocom_2}
\end{figure*}

\begin{figure*}[!ht]
\centering{\includegraphics[width=\linewidth]{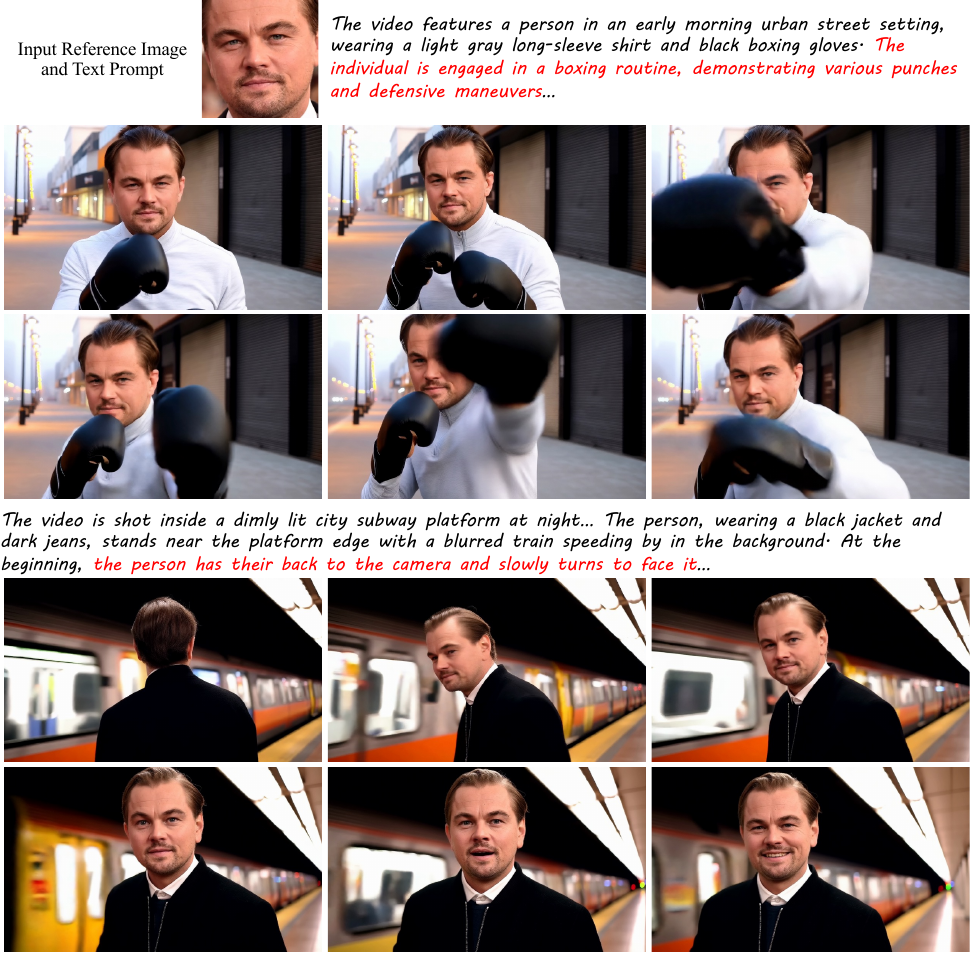}}
	\caption{More showcases of identity-preserving videos generated by our FaithfulFaces.
	}
	\label{fig:append_vis_nocom_3}
\end{figure*}

\begin{figure*}[!ht]
\centering{\includegraphics[width=\linewidth]{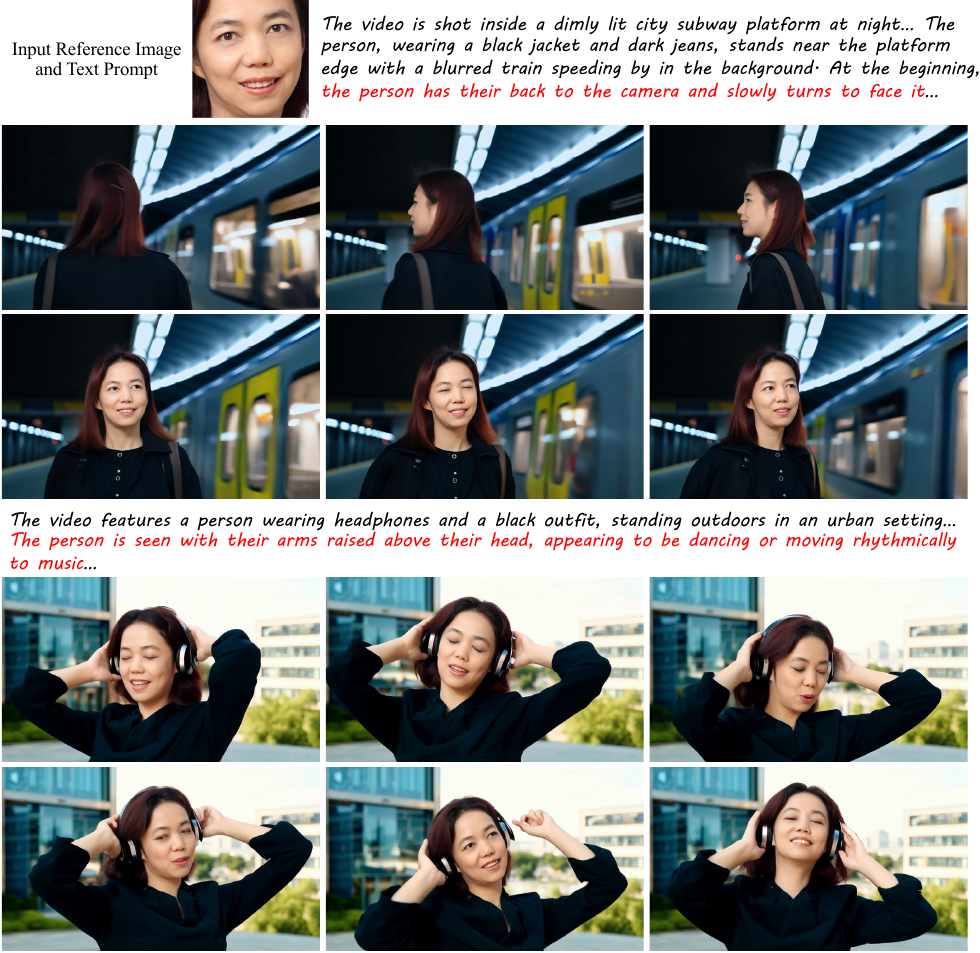}}
	\caption{More showcases of identity-preserving videos generated by our FaithfulFaces.
	}
	\label{fig:append_vis_nocom_4}
\end{figure*}

In this section, we provide more visual comparisons of different methods in Figs.~\ref{fig:append_vis_1},~\ref{fig:append_vis_2},~\ref{fig:vis_2}, and~\ref{fig:append_vis_3} to demonstrate the effectiveness of our method.
Additionally, we provide more showcases of identity-preserving videos generated by our FaithfulFaces in Figs.~\ref{fig:append_vis_nocom_1},~\ref{fig:append_vis_nocom_2},~\ref{fig:append_vis_nocom_3}, and~\ref{fig:append_vis_nocom_4}, covering a variety of identities, actions, and scenes.

\clearpage

\clearpage

\clearpage

\clearpage

\clearpage

\clearpage

\clearpage

\clearpage


\newpage
\section*{NeurIPS Paper Checklist}

\begin{enumerate}

\item {\bf Claims}
    \item[] Question: Do the main claims made in the abstract and introduction accurately reflect the paper's contributions and scope?
    \item[] Answer: \answerYes{} 
    \item[] Justification: The key contributions claimed in the abstract and introduction are elaborated in Sec.~\ref{gfprl} and Sec.~\ref{pipeline}. We provide extensive quantitative and qualitative experimental results to verify the effectiveness of these contributions.

\item {\bf Limitations}
    \item[] Question: Does the paper discuss the limitations of the work performed by the authors?
    \item[] Answer: \answerYes{} 
    \item[] Justification: The limitations are discussed in Sec. 5, Conclusion.

\item {\bf Theory assumptions and proofs}
    \item[] Question: For each theoretical result, does the paper provide the full set of assumptions and a complete (and correct) proof?
    \item[] Answer: \answerNA{} 
    \item[] Justification: This paper does not include theoretical results.

    \item {\bf Experimental result reproducibility}
    \item[] Question: Does the paper fully disclose all the information needed to reproduce the main experimental results of the paper to the extent that it affects the main claims and/or conclusions of the paper (regardless of whether the code and data are provided or not)?
    \item[] Answer: \answerYes{} 
    \item[] Justification: First, we have explained the implementation of  FaithfulFaces in detail in Sec.~\ref{Implementation}. Second, we have explained the details of training and inference in Fig.~\ref{fig:2} and Sec.~\ref{Overview}. Third, we have explained the details of the dataset construction in Sec.~\ref{pipeline}. In addition, the codes will be publicly available when the paper is accepted.

\item {\bf Open access to data and code}
    \item[] Question: Does the paper provide open access to the data and code, with sufficient instructions to faithfully reproduce the main experimental results, as described in supplemental material?
    \item[] Answer: \answerNo{} 
    \item[] Justification: No code is included in this submission.

\item {\bf Experimental setting/details}
    \item[] Question: Does the paper specify all the training and test details (e.g., data splits, hyperparameters, how they were chosen, type of optimizer) necessary to understand the results?
    \item[] Answer: \answerYes{} 
    \item[] Justification: We provide the implementation details in Sec.~\ref{Implementation}.

\item {\bf Experiment statistical significance}
    \item[] Question: Does the paper report error bars suitably and correctly defined or other appropriate information about the statistical significance of the experiments?
    \item[] Answer: \answerYes{} 
    \item[] Justification: While standard error bars are computationally prohibitive for large-scale video generation tasks, we ensure statistical significance by evaluating our method across a comprehensive test set of 600 diverse video generations. We use robust, standardized metrics (FaceSim, FID, CLIPScore) over this entire set to reliably demonstrate our performance gains.

\item {\bf Experiments compute resources}
    \item[] Question: For each experiment, does the paper provide sufficient information on the computer resources (type of compute workers, memory, time of execution) needed to reproduce the experiments?
    \item[] Answer: \answerYes{} 
    \item[] Justification: Sec.~\ref{Implementation} provides the type of compute workers GPU.
    
\item {\bf Code of ethics}
    \item[] Question: Does the research conducted in the paper conform, in every respect, with the NeurIPS Code of Ethics \url{https://neurips.cc/public/EthicsGuidelines}?
    \item[] Answer: \answerYes{} 
    \item[] Justification: We have reviewed the NeurIPS Code of Ethics.

\item {\bf Broader impacts}
    \item[] Question: Does the paper discuss both potential positive societal impacts and negative societal impacts of the work performed?
    \item[] Answer: \answerYes{} 
    \item[] Justification: We discuss the Broader Impact in Appendix~\ref{Ethics}.
    
\item {\bf Safeguards}
    \item[] Question: Does the paper describe safeguards that have been put in place for responsible release of data or models that have a high risk for misuse (e.g., pre-trained language models, image generators, or scraped datasets)?
    \item[] Answer: \answerNA{} 
    \item[] Justification: We do not publicly release the training dataset or trained model weights, nor do we provide a public inference API. Potential misuse risks are discussed in Appendix A.15.

\item {\bf Licenses for existing assets}
    \item[] Question: Are the creators or original owners of assets (e.g., code, data, models), used in the paper, properly credited and are the license and terms of use explicitly mentioned and properly respected?
    \item[] Answer: \answerYes{} 
    \item[] Justification: All assets used in this paper are properly credited in Sec.~\ref{Implementation}.

\item {\bf New assets}
    \item[] Question: Are new assets introduced in the paper well documented and is the documentation provided alongside the assets?
    \item[] Answer: \answerNA{} 
    \item[] Justification: This paper does not release new assets.

\item {\bf Crowdsourcing and research with human subjects}
    \item[] Question: For crowdsourcing experiments and research with human subjects, does the paper include the full text of instructions given to participants and screenshots, if applicable, as well as details about compensation (if any)? 
    \item[] Answer: \answerNA{} 
    \item[] Justification: This paper does not involve crowdsourcing nor research with human subjects.

\item {\bf Institutional review board (IRB) approvals or equivalent for research with human subjects}
    \item[] Question: Does the paper describe potential risks incurred by study participants, whether such risks were disclosed to the subjects, and whether Institutional Review Board (IRB) approvals (or an equivalent approval/review based on the requirements of your country or institution) were obtained?
    \item[] Answer: \answerNA{} 
    \item[] Justification: This paper does not involve crowdsourcing nor research with human subjects.

\item {\bf Declaration of LLM usage}
    \item[] Question: Does the paper describe the usage of LLMs if it is an important, original, or non-standard component of the core methods in this research? Note that if the LLM is used only for writing, editing, or formatting purposes and does \emph{not} impact the core methodology, scientific rigor, or originality of the research, declaration is not required.
    \item[] Answer: \answerNA{} 
    \item[] Justification: This submission only utilizes LLMs for grammar checking.

\end{enumerate}

\end{document}